\documentclass[pdflatex,sn-basic]{sn-jnl}

\usepackage{graphicx}%
\usepackage{multirow}%
\usepackage{amsmath,amssymb,amsfonts}%
\usepackage{amsthm}%
\usepackage{mathrsfs}%
\usepackage[title]{appendix}%
\usepackage{xcolor}%
\usepackage{textcomp}%
\usepackage{manyfoot}%
\usepackage{booktabs}%
\usepackage{algorithm}%
\usepackage{algorithmicx}%
\usepackage{algpseudocode}%
\usepackage{listings}%

\usepackage{adjustbox} 
\usepackage{colortbl} 
\usepackage{caption} 
\usepackage{subcaption} 
\usepackage{verbatim} 

\definecolor{lightgray}{gray}{0.8} 

\begin{document}

\title[]{Learning with Confidence: Training Better Classifiers from Soft Labels}

\author*[1,2]{\fnm{Sjoerd} \spfx{de} \sur{Vries}}\email{s.devries1@uu.nl}

\author[1]{\fnm{Dirk} \sur{Thierens}}

\affil*[1]{\orgdiv{Department of Information and Computing Sciences}, \orgname{ Utrecht University}, \orgaddress{ \street{Princetonplein 5}, \postcode{3584 CC}, \city{Utrecht}, \country{The Netherlands}}} 

\affil[2]{\orgdiv{Department of Digital Health}, \orgname{University Medical Center Utrecht}, \orgaddress{ \street{Heidelberglaan 100}, \postcode{3584 CX}, \city{Utrecht}, \country{The Netherlands}}}  

\abstract{In supervised machine learning, models are typically trained using data with hard labels, i.e., definite assignments of class membership. This traditional approach, however, does not take the inherent uncertainty in these labels into account. We investigate whether incorporating label uncertainty, represented as discrete probability distributions over the class labels---known as soft labels---improves the predictive performance of classification models. We first demonstrate the potential value of soft label learning (SLL) for estimating model parameters in a simulation experiment, particularly for limited sample sizes and imbalanced data. Subsequently, we compare the performance of various wrapper methods for learning from both hard and soft labels using identical base classifiers. On real-world-inspired synthetic data with clean labels, the SLL methods consistently outperform hard label methods. Since real-world data is often noisy and precise soft labels are challenging to obtain, we study the effect that noisy probability estimates have on model performance. Alongside conventional noise models, our study examines four types of miscalibration that are known to affect human annotators. The results show that SLL methods outperform the hard label methods in the majority of settings. Finally, we evaluate the methods on a real-world dataset with confidence scores, where the SLL methods are shown to match the traditional methods for predicting the (noisy) hard labels while providing more accurate confidence estimates.}


\maketitle

\section{Introduction}\label{sec:intro}

In the field of machine learning, the development of prediction models for classifying data into their respective classes has received substantial attention. The effectiveness of these models largely depends on the availability of labelled data to enable supervised training. However, obtaining accurate labels often proves to be difficult and may require considerable human effort.

Typically, data annotation is carried out by either a small group of experts or a larger number of unskilled workers, known as crowdsourcing. For instance, in healthcare, domain-specific expertise from physicians is often required to obtain labels of sufficient quality. Despite their expertise, such professionals may not have all the necessary information available to them, nor perfect knowledge of the task, resulting in uncertainty in their annotations. Similarly, multiple unskilled workers might assign different labels to an instance, causing uncertainty.

While labelling a single example can be time-consuming, particularly for complex problems, estimating the uncertainty associated with a label---such as through a confidence score---typically requires minimal additional effort~\citep{nguyen2014learning, song2018active}. When multiple annotators, whether they are experts or crowdsourcing workers, label the same data point, the process yields multiple labels. In practice, these confidences or multiple labels are often consolidated into a single, definitive class assignment, as most classification methods are designed to work with such hard labels. For example, data about which an expert is uncertain may be reviewed by colleagues to obtain a consensus label, after which the confidence score is discarded since it cannot be used as predictive feature. Similarly, multiple label sets are often aggregated through methods such as plurality voting, causing the available information about the uncertainty to be lost. 

Methods that take uncertainty into account are being researched, especially in the context of crowdsourcing~\citep{zhang2022knowledge}. However, methods from this field are often tailored to a specific setting, assuming for example that information about which annotator generated a label is available. Training a model directly using confidence scores or other measures of uncertainty remains underexplored. While binary classification tasks can be tackled with regression models or by weighting examples, multi-class classification is more challenging, with few universally applicable methods available. We believe this is a missed opportunity, as uncertainty estimates have the potential to provide valuable information for training classification models~\citep{raykar2010learning, zhang2018ensemble}.

Given that the more specific a method is, the less likely it is to achieve widespread adoption, we generalize learning with confidences to the setting of soft label learning (SLL). Soft labels are defined as discrete probability distributions over the class labels, representing the probabilities that an instance belongs to the corresponding classes. Problems from various fields can be reformulated into an SLL problem, thereby broadening the applicability of SLL methods beyond learning from confidence scores. The central question we investigate is whether SLL can be leveraged to train classification models that effectively take uncertainty into account, leading to improved performance compared to models trained with hard-labelled data. 

Moreover, soft labels may be subject to uncertainty themselves as a consequence of human estimation error, i.e. miscalibration. The impact of such label noise, and particularly on soft labels, has not received much attention. Studies that address SLL in combination with noise, often limit their scope to Gaussian noise~\citep{peng2014learning, xue2017efficient}, which might not realistically represent the noise typically introduced by human annotators. In this work we therefore direct our attention toward a different type of noise: miscalibration.

We implement a variety of techniques and methods for learning from soft labels and develop new methods by integrating these techniques in various ways. The methods we investigate are wrapper methods that can be used with any classifier that can provide probability estimates and handle weights. This ensures that performance comparisons between soft- and hard-labelled methods are fair, as they are tested using the same base classifiers. We anticipate that ensemble learning will be especially effective, as ensembles benefit from diversity~\citep{brown2005diversity} and might therefore be better equipped to deal with any noise due to miscalibration. Our experiments are conducted on simulated, realistic-synthetic and a real-world dataset that has expert assigned confidence scores: the UrinCheck dataset for predicting urinary tract infection in hospital patients~\cite{de2022semi}. Additionally, we analyse how both types of methods perform under label noise introduced both through traditional noise models and miscalibration.

The primary contributions of our work are as follows:

\begin{itemize}
\item A simulation study that demonstrates the potential of learning from soft labels to improve classification models.
\item The implementation of existing and introduction of several new SLL wrapper methods that facilitate a fair comparison between SLL and traditional supervised learning.
\item The inclusion of four miscalibration noise models specific to soft labels, studied alongside more commonly used noise models.
\item A comprehensive comparison of the various methods, both on realistic synthetic data as well as real-world data, showing that SLL methods can effectively leverage label uncertainty to achieve superior performance compared to methods that do not incorporate this information.
\end{itemize}

\section{Related Work}

Several studies have explored learning from data with confidence scores, the same setting as that of the UrinCheck dataset for predicting urinary tract infections (UTI) that we discuss in Section~\ref{sec:urincheck}. One of the early works in this domain is by~\cite{oyama2013accurate}, where the authors leverage self-reported confidence scores from multiple annotators. They use the expectation maximization (EM) based Dawid-Skene method~\citep{dawid1979maximum} to iteratively estimate both the accuracies of the crowd workers and the final labels, leading to enhanced accuracy of the integrated labels. Similarly, ~\cite{reamaroon2018accounting} address a problem that closely resembles the UTI prediction task: training a model on a medical dataset where physicians quantify their diagnostic uncertainty. Their method involves adapting a support vector machine (SVM) to handle the label uncertainty, although their method focuses on longitudinal data. Another relevant work in the medical domain investigates heparin induced thrombocytopenia~\citep{nguyen2014learning}, where experts provided both labels and confidence scores for each instance. The authors adapt regression methods to incorporate the confidence scores and employ ranking to improve the noise tolerance of an SVM for binary classification. Their methods demonstrate improvements in AUC, especially when few samples were available.

Further studies into learning from confidence scores can be found in the label noise literature. Most often label noise models are investigated that are simplistic in that they allow for noise which is Noisy Completely at Random (NCAR) or depends on the label, Noisy at Random (NAR)~\citep{frenay2013classification}. However, when individual estimates of the noise level are made and the noise is Noisy Not at Random (NNAR), these estimates can be treated like confidence scores. For instance,~\cite{gui2015novel} propose a k-nearest neighbours approach to estimate individual noise levels, followed by two methods for binary classification that incorporate these confidences: modifying a surrogate loss function and sampling based on confidence scores. These approaches outperform other methods when dealing with noisy data, and we employ the latter technique in some of the methods tested in our study. ~\cite{berthon2021confidence} introduce learning with confidence-scored instance-dependent noise. They also utilise a confidence score to quantify the noise level of an individual data point. Their instance-level forward correction algorithm iteratively updates an instance-dependent transition matrix, outperforming existing methods and working for multi-class problems. 

A shared limitation of the aforementioned methods is their reliance on confidence scores or individual noise rate estimates for each instance. This approach only considers information about the most probable class, neglecting any further information about the other classes in multi-class problems. In this work investigate methods that can be applied not only to data with confidence scores but also to a broader class of problems: soft label learning (SLL). Methods that handle soft labels are versatile in that they can be applied to learning with confidences, as well as tasks where multiple expert or crowd annotators have provided labels for a particular instance, or probability estimates were provided by a prediction model or an annotator directly.

In the field of learning from crowds or multiple annotators, a single instance is labelled repeatedly, resulting in a set of labels for each instance. While such a set can be transformed into a soft label, the annotations are frequently integrated into a hard label instead through a process called ground truth inference~\citep{zhang2016learning}, which risks losing information about the inherent uncertainty of the instance. Other methods do take the individual annotation into account explicitly. For example, ~\cite{raykar2010learning} propose an algorithm that jointly learns a prediction model, the expertise level of the annotators and the true label set. Such joint optimisation, based on the EM algorithm, is frequently used in crowd labelling~\citep{zhang2016learning}, and was also employed in the aforementioned work by~\cite{oyama2013accurate}. However, these algorithms require additional information, such as which annotator assigned which label, limiting their applicability. To ensure the methods in this work are as general as possible, we assume that no additional information is available beyond the soft labels themselves, disqualifying the previous methods that take into account individual worker performance. 

One of the earliest works to address learning from soft labels directly is~\cite{jin2002learning}, who refer to this task as the ``multiple-label" problem. They iteratively fit a conditional model and the label distribution using EM. Other works focus on belief functions to handle the probability distributions over the class space by experts~\citep{denoeux2001handling, come2009learning}. In~\cite{geng2016label} a different but related problem is introduced: label distribution learning (LDL).  In this task, and instance can have multiple true classes, each to a different extent. This is also captured in a soft label, allowing for the possibility of transferring algorithms between the LDL and SLL. ~\cite{gao2017deep} utilise soft labels to minimize the Kullback-Leibler divergence between predicted and ground truth labels for neural networks, effectively reducing overfitting. ~\cite{peterson2019human} further explore learning neural networks directly from soft labels, resulting in improved generalization. A more indirect use of soft labels is proposed by~\cite{fornaciari2021beyond}, where learning the soft labels is presented as an auxiliary task to mitigate overfitting. 

Although these methods learn directly from soft labels, they are highly specialized and depend on customization of the loss function. For potential users, researching, selecting, implementing and comparing multiple such methods with more conventional approaches can be impractical. Our primary interest lies in studying methods that can be used in conjunction with common base classifiers, such that for any particular problem their added value can be tested with minimal additional effort required, i.e. wrapper methods. Furthermore, the use of the same base classifiers in our experiments makes for a fair comparison between hard- and soft label methods.

Such methods are scarce in the literature, however. In their works, Sheng et al focus on binary problems, either employing methods for duplicating data or weighting examples~\citep{sheng2008get, sheng2011simple}. Another work which describes a method for that can be applied to SLL is~\cite{zhang2018ensemble}. They first create different datasets by bootstrapping, then duplicate instances with multiple labels and weight them according to their label frequencies.   

In summary, our research focusses on generally applicable SLL methods. We believe that how SLL methods handle different types of noise is crucial to their real-world performance. In~\cite{xue2017efficient}, the authors build on their previous work where they study SLL~\citep{nguyen2011learning}, by emphasising the influence of Gaussian noise (NCAR) on the soft labels, finding that binning helps to reduce the impact of such noise. ~\cite{peng2014learning} also address the problem of noise in probabilistic labels. They develop the Fractional Score-based Classifier based on Gaussian Process Regression and experiment with Gaussian noise. However, these approaches are not wrapper methods.

In this work, we conduct a thorough investigation of how SLL methods perform compared to hard label learning (HLL) methods. We direct considerable effort to studying the effects of various types of label noise, including noise models that are known to affect human probability estimates from the psychology literature, as detailed in Section~\ref{sec:noise_types}.

\section{Advantage of Soft Labels: Simulated Data}\label{sec:gaussians}

We initiate our study by exploring the potential benefits of using well-calibrated soft label through an experiment with simulated data. The question we seek to answer with this experiment is whether learning from soft labels rather than hard labels can lead to improved models. We define improved to mean that the model parameter estimates based on the soft labels are closer the values of the ground truth model that was used to generate the data, than the parameter estimates obtained based on the hard labels.

\subsection{Clean Gaussian Data}\label{sec:gaussian_noiseless}

\begin{figure}[!t]
    \centering
    \begin{subfigure}[b]{0.495\textwidth}
        \centering
        \includegraphics[width=\textwidth, quiet]{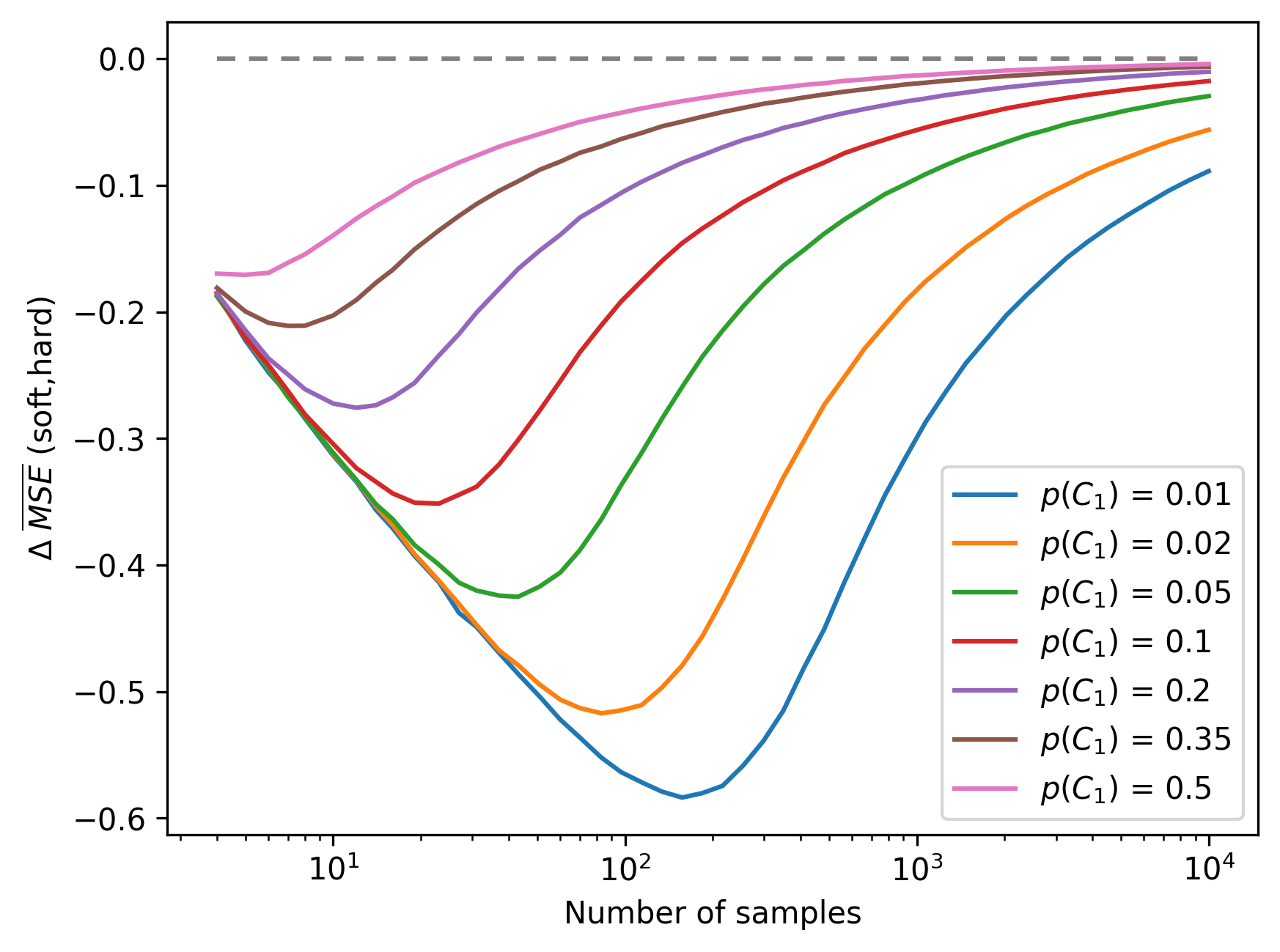}
        \caption{Without noise ($\sigma = 0.0$)}
        \label{fig:gaussian_noiseless}
    \end{subfigure}
    \hfill
    \begin{subfigure}[b]{0.495\textwidth}
        \centering
        \includegraphics[width=\textwidth, quiet]{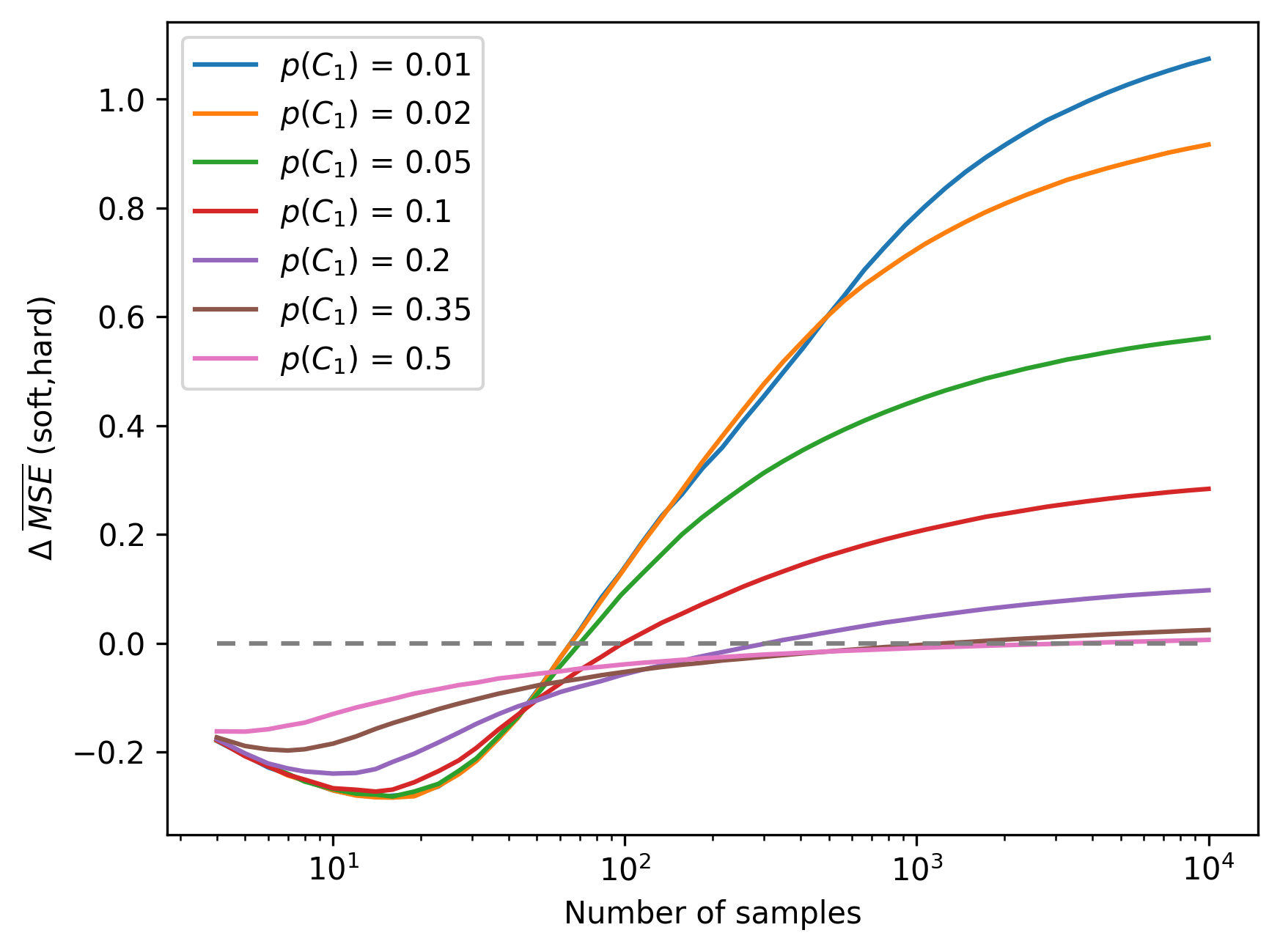}
        \caption{With noise ($\sigma = 0.1$)}
        \label{fig:gaussian_with_noise}
    \end{subfigure}
    \caption{$\Delta \overline{MSE}$ (soft,hard) for different number of samples taken from the true class distributions for different values of the prior probability of class one, $p(C_1)$. Shown without noise (a) and with noise (b) added to the soft labels.}
    \label{fig:mainfig}
\end{figure}

In this experiment, we generate data by sampling from two multivariate normal distributions $MVN_1$ and $MVN_2$, representing class one $(C_1)$ and class two $(C_2)$ respectively. These distributions are symmetrically positioned with respect to the origin, with both distributions having the identity matrix as their covariance. We set the number of problem dimensions $n$ equal to two, and the distance parameter $d$, which is used to determine the mean of the distributions as $\mu_1 = (d)^n$ and $\mu_2 = (-d)^n$ respectively, to be 0.5. We vary the the imbalance of the dataset, i.e. the prior probability $p(C_1)$ that an instance is sampled from the $MVN_1$, from 0.01 to 0.5. This fixes the probability $p(C_2)$ that an instance is sampled from $MVN_2$ to be $1-p(C_1)$. 

For each imbalance setting, we vary the number of samples drawn from the distributions from 4 to 10.000 samples, ensuring that there are always at least two samples of each class included. From these samples, we estimate the mean of each distribution. 

To obtain the hard label estimates of the mean for each class, $\hat{\mu}_{1,hard}$ and $\hat{\mu}_{2,hard}$, we calculate the unweighted average over the values in each dimension across all samples belonging to either class. 

To generate a soft label for each instance, we have to estimate the probability that a sample $x$ to belongs to either class one, $p_{C_1}(x)$, or class two, $p_{C_2}(x)$. This can be achieved by applying the following formula:

\begin{equation}\label{eq:gaussian_probability}
p_{C_1}(x) = \frac{p(C_1) \cdot p_{MVN_1}(x)}{p(C_1) \cdot p_{MVN_1}(x) + p(C_2)\cdot p_{MVN_2}(x)},
\end{equation}

where $p_{MVN_1}$ and $p_{MVN_2}$ are the probability density functions of $MVN_1$ and $MVN_2$ respectively, given sample $x$. $p_{C_2}$ is defined similarly. Together they form the soft label $[p_{C_1}, p_{C_2}]$ belonging to $x$. To get the soft label estimates of the mean for each class, $\hat{\mu}_{1,soft}$ and $\hat{\mu}_{2,soft}$, we then take the weighted average over all samples, using the probabilities $p_{C_1}$ and $p_{C_2}$ as weights. The mean squared error $MSE$ between the true model mean for class one $\mu_1$, and the estimated means, $\hat{\mu}_{1,soft}$ and $\hat{\mu}_{1,hard}$ is then calculated and averaged over 100.000 repeats of the procedure for each sample size.

The results for different $p(C_1)$ are presented in Figure~\ref{fig:gaussian_noiseless}. The figure shows $\Delta \overline{MSE}$ (soft,hard), defined as the average $MSE(\hat{\mu}_{1,soft}, \mu_1)$ minus the average $MSE(\hat{\mu}_{1,hard},\mu_1)$. A negative $\Delta \overline{MSE}$ (soft,hard) thus indicates that on average the value of $\hat{\mu}_{1,soft}$ is closer to $\mu_1$ than $\hat{\mu}_{1,hard}$ is. We observe that this holds true for all sample sizes, leading us to conclude that soft labels can be used successfully for learning more accurate model parameters.

We observe that the difference is largest for small sample sizes. This can be explained by the fact that the information that is contained in a single soft label [$p_{C_1}$, $p_{C_2}$] can only be expressed by combining multiple hard labels. For instance, a data point with soft label [0.75,0.25] at position (0.25, 0.25) may be approximated by three hard-labelled points of $C_1$: [1, 0] and one point of $C_2$: [0, 1], with (0.25, 0.25) at the same location. In essence, a soft label can convey more information per data point than a hard label.

Additionally, the distinction between soft and hard label performance increases with greater class imbalance. The reasoning behind this is similar: under high class imbalance, there are fewer data points available for one class and the information describing that class is better captured by the soft labels. The smaller differences observed at the really small number of samples can be attributed to the requirement that each sample contains at least two points from each class, effectively creating a balanced scenario. Only when the sample size increases, the imbalance is actually introduced into the data.

\subsection{Random Noise}\label{sec:gaussian_noise}

The previous results pertain to a scenario in which the soft labels are perfectly accurate. However, in real-world applications, both hard and soft labels are often subject to noise. We hypothesise that as noise is added to the soft labels, their added value will decrease.

To test this hypothesis, we repeat the experiment from the previous subsection, using the same values for $d$, $n$ and $p(C_1)$, while introducing Gaussian noise to the soft labels. It is important to note that the hard labels remain noiseless, allowing us to determine the number of samples for which it becomes advantageous to have clean hard labels, rather than noisy soft labels. Specifically, we add noise by sampling a value from a normal distribution with $\mu=0.0$ and $\sigma = 0.1$ and adding it to $p_{C_1}$, while subtracting it from $p_{C_2}$, as calculated using Equation~\ref{eq:gaussian_probability}, and truncating the values to ensure they fall with the interval from 0 to 1.

The results for estimating $\mu_1$ for different prior probabilities $p(C_1)$ are shown in Figure~\ref{fig:gaussian_with_noise}. As anticipated, we observe that the soft labels become less useful as noise is introduced. While in the previous experiment the difference between $\hat{\mu}_{1,soft}$ and $\hat{\mu}_{1,hard}$ converges to zero as the number of samples increases, the addition of noise to the soft labels causes the hard labels to be more valuable beyond a certain number of samples for all of the data imbalance scenarios. Notably, at approximately 50 samples we observe similar performance for all imbalance settings and for larger sample sizes the more imbalanced scenarios result in a larger advantage of learning from hard rather than soft labels. 

This behaviour can be explained as follows: when sufficient data is available, the information contained in a single soft label can be effectively approximated by multiple hard labels. When the soft labels are noisy while the hard labels are not, the hard labels become more informative as sample size increases. In situations with limited samples, a noisy soft label may still offer an advantage over a hard label. Although the information contained in the soft label may not be entirely correct, in sparse areas of the feature space there might not be enough hard-labelled instances to convey the necessary information about the uncertainty at that location.

This simple experiment demonstrates that learning from soft labels can be beneficial, especially for small sample sizes. Moreover, this advantage persisted when data were scarce even when noise was introduced  exclusively to the soft labels. However, as the sample size increased, noise caused the soft labels to become less informative.

\section{Methods}

In this section, we present the methods and data generation processes used in our experiments. Section~\ref{sec:techniques} outlines several techniques, which are combined to form methods for learning from soft labels. In Section~\ref{sec:synlabel_data} we detail how realistic synthetic data with both clean hard and soft labels is generated. Finally, in Section~\ref{sec:noise_types} we present the noise models studied in this work, including four distinct types of miscalibration that can be applied to soft labels. 

\subsection{Soft Label Learning}\label{sec:techniques}

Problems that can be framed as soft label learning (SLL) arise in several fields, including crowdsourcing and learning from expert annotators, model outputs or confidence assessments. In this work, we investigate methods that enable a classifier to learn from soft labels, regardless of their origin. Many existing approaches that handle multiple labels per instance are tailored to specific scenarios, such as when the quality of the annotators can be modelled. We address the more general SLL setting, where only the integrated soft label and feature information are available. Because this setting makes minimal assumptions about the label generation process, SLL methods are are broadly applicable. 

Additionally, the methods we study are wrapper methods, designed to work with any base classifier capable of producing probability estimates and taking weights into account. This allows us to measure the performance of various SLL and hard label learning (HLL) methods using the same classifiers, ensuring a fair comparison.

\subsubsection{Algorithms}\label{sec:algorithms}

\begin{table*}[!t]
    \centering
    \begin{adjustbox}{max width=\textwidth}
    \begin{tabular}{llccccc} 
        \hline
        Method Type & Method & Threshold & PV & Instance Sampling & Duplication & Soft Label Processing \\
        \hline
        \multirow[c]{5}{*}{HardSingle} & PluralityClassifier & - & X & - & - & - \\ 
        & PluralityWeightsClassifier & - & X & - & - & Weights \\ 
        & ThresholdClassifier & X & X & - & - & - \\ 
        & ThresholdWeightsClassifier & X & X & - & - & Weights \\ 
        \arrayrulecolor{lightgray}\hline
		\multirow[c]{2}{*}{SoftSingle} & SampleClassifier & - & - & - & - & Label Sampling \\
        
        & DuplicateWeightsClassifier & - & - & - & X & Weights \\
        \arrayrulecolor{lightgray}\hline
        \multirow[c]{6}{*}{HardEns} & PluralityBootstrapClassifier & - & X & Bootstrap & - & - \\ 
        & PluralityBootstrapWeightsClassifier & - & X & Bootstrap & - & Weights \\ 
        & PluralityEnsembleClassifier & - & X & Max Sampling & - & - \\ 
        & ThresholdBootstrapClassifier & X & X & Bootstrap & - & - \\ 
        & ThresholdBootstrapWeightsClassifier & X & X & Bootstrap & - & Weights \\ 
        & ThresholdEnsembleClassifier & X & X & Max Sampling & - & - \\ 
        \arrayrulecolor{lightgray}\hline
        \multirow[c]{7}{*}{SoftEns} & BootstrapSamplingClassifier & - & - & Bootstrap & - & Label Sampling \\ 
        & EnsembleSamplingClassifier & - & - & Max Sampling & - & Label Sampling \\ 
        & DuplicateEnsembleClassifier & - & - & - & X & Max Sampling \\
        & BootstrapDuplicateWeightsClassifier & - & - & Bootstrap & X & Weights \\ 
        & EnsembleDuplicateWeightsClassifier & - & - & Max Sampling & X & Weights \\
        & BootstrapDuplicateSamplingClassifier & - & - & Bootstrap & X & Max Sampling \\
        & EnsembleDuplicateSamplingClassifier & - & - & Max Sampling & X & Max Sampling \\
        
        \hline
    \end{tabular}
    \end{adjustbox}
    \caption{The methods studied in this work, along with the techniques they consist of. Method names are abbreviated throughout this study: Classifier - Clf, Duplicate - Dup, Ensemble - Ens, WeightedFit - WF. For example, EnsembleDuplicateWeightsClassifier is abbreviated as EnsDupWeightsClf. The order in which the techniques are applied follows the column order: first Thresholding is applied, then Plurality Voting (PV), then Instance Sampling, then Duplication and finally any further Soft Label Processing techniques.}
    \label{tab:methods}
\end{table*}

In the literature, several wrapper methods have been proposed for SLL. These methods employ a variety of techniques, including the following:

\vspace{8pt}
\textbf{Thresholding} involves setting a threshold, used to exclude a percentage of the least confident instances from the training data. Alternatively, a hard probability bound can be set for the plurality class, discarding any instances that have a lower maximum probability value. The labels remains soft labels after applying this technique, requiring the application of a technique methods such as plurality voting before most classifiers can be trained on these data.

\vspace{8pt}
\textbf{Plurality Voting (PV)} selects the class with the highest probability as the hard label. This one of the most straightforward approaches for handling soft labels. In ensemble learning, PV refers to selecting the class that is predicted by most ensemble members for a certain instance as the ensemble prediction. For binary problems, PV is equivalent to majority voting (MV).

\vspace{8pt}
\textbf{Instance Sampling} involves creating an ensemble by sampling instances from the original dataset to form a new training set for each ensemble member. In this study, we use either bootstrap sampling (bagging~\citep{breiman1996bagging}) or max sampling. In bagging, the soft labels are not considered, while in max sampling, instances are sampled with replacement based on the highest class probability in the soft label. This can be interpreted as sampling according to the confidence that the label is correct, i.e. 1 - the individual error rate, as in~\cite{gui2015novel}.  

\vspace{8pt}
\textbf{Duplication} is a technique where each example is duplicated into $k$ instances, with $k$ the number of classes. The original probabilities in the soft label for these classes are typically used as weights in a subsequent step, ensuring that the uncertainty information of the original example is preserved. Alternatively, a single example with multiple annotations can be split into as many separate data points, each with one associated hard label~\citep{raykar2010learning}. In this work, the former approach is studied as the origin of the soft label is assumed to be unknown.

\vspace{8pt}
\textbf{Weighting} is an option available in some classifiers that enables them to account for the weight of an example during training. When combined with duplication, this allows for the conversion of a dataset with soft labels into a larger dataset of weighted examples with hard labels, on which a classifier can be trained. When combined with label sampling or PV, techniques that assign a hard label to the resulting instances, weighting helps preserve some of the information from the original soft label.

\vspace{8pt}
\textbf{Label Sampling} involves sampling a hard label from a soft label based on the probabilities corresponding to the classes in the soft label.

\vspace{8pt}
\textbf{Ensemble Learning} entails creating multiple classifiers, each producing a prediction that is combined into a final ensemble prediction. In this work, these ensembles are created by through instance sampling. Furthermore, to arrive at en ensemble prediction, the probability estimates for each instance are averaged over all ensemble members.   

\vspace{8pt}

These techniques are employed by different methods found in the literature, which we describe here using their constituent techniques for improved clarity. ~\cite{sheng2008get} proposed the Multiplied Examples approach for SLL, which uses duplication followed by weighting to learn from soft labels, referred to here as the DuplicateWeightsClassifier. They compared this method with simply applying MV, which corresponds to PV for multi-class problems, to multiple label sets to obtain a single hard label for each instance, referred to here as the PluralityClassifier.

Sheng then extended both of these binary classification methods~\citep{sheng2011simple}. The MV approach was expanded with the MV-freq methods, where weighting was used to include additional information about the uncertainty of the majority class, referred to here as the PluralityWeightsClassifier. The Multiplied Examples approach was named the pairwise approach, or Paired-Freq. Additionally they introduced a weighting scheme, specifically tailored to binary problems. The MV approaches are compared to the duplication based approaches and showed the latter to have superior performance.

Sampling, as proposed in~\cite{zadrozny2003cost}, offers an alternative to weighting for incorporating soft label information. By applying max sampling after duplication, instance-label pairs can be selected for inclusion in the subsample that an ensemble member is trained on. The DuplicateEnsembleClassifier, Bootstrap- and EnsembleDuplicateSamplingClassifiers employ this technique. Alternatively, a definitive label can be selected for a specific instance using label sampling, as in the SampleClassifier and Bootstrap- and EnsembleSamplingClassifiers. 

In \cite{zhang2018ensemble} the EnsembleMV method is proposed, which combines the techniques of bootstrapping, duplication and weighting to learn from soft-labelled data. In this work, we refer to this classifier the BootstrapDuplicateWeightsClassifier to highlight the techniques it incorporates. Additionally, they use MVBagging (referred to here as the PluralityBootstrapClassifier) as a baseline comparison method, which applies PV followed by Bagging.

\vspace{8pt}
In addition to the methods from the literature, we developed new SLL methods by combining the discussed techniques. The resulting methods, along with their constituent techniques, are presented in Table~\ref{tab:methods}. These methods are primarily distinguished along two axes: (1) whether they use information from the entire soft label (SLL) or from at most one of the classes (HLL), and (2) whether ensemble learning is employed. We organise the methods according to these axes:

\begin{itemize}
\item \textbf{HardSingle} methods, that only consider the probability estimate of the most likely class in the soft label when training a model, treating it as a (weighted) hard label, and use a single classifier.
\item \textbf{SoftSingle} methods, that consider the probability estimates of all of the classes in the soft labels, and use a single classifier.
\item \textbf{HardEns} methods, that only consider the probability estimate of the most likely class in the soft label when training a model, and employ multiple classifiers to form an ensemble.
\item \textbf{SoftEns} methods, that take into account all of the information contained in the soft labels and construct an ensemble of classifiers to make a final prediction.
\end{itemize}

Throughout this work, we compare these categories to assess which type of methods is the most effective.

\subsubsection{Base Classifiers}

We expect that the behaviour of the different SLL and HLL methods will vary depending on the base classifier that they are used with. To derive general conclusions about these method, we employ a variety of base classifier in our experiments. Since one of the techniques used in these methods is weighting, we selected classifiers that natively support weights, could output probability estimates and were implemented in scikit-learn~\citep{scikit-learn}: Decision Tree (DT), or a Random Forest (RF) when used in a bagged ensemble, Logistic Regression (LR), a quadratically smoothed Support Vector Machine (SVM) with $\gamma = 2$ , implemented via the Stochastic Gradient Descent (SGD) classifier with modified Huber loss, and finally Gaussian Naive Bayes (GNB).

We mostly adhere to the default setting, the main parameters and their used values are listed in Table~\ref{tab:classifiers}.  We opted for the SGD classifier over other SVM implementations due to its speed and its ability to produce probability estimates. 

\subsection{Obtaining Soft-labelled Data}\label{sec:synlabel_data}

To effectively evaluate the methods, ideally both clean data for performance measurements and noisy data that represent a real-world learning scenario for training are available. Furthermore, both types of datasets are required to include both hard and soft labels for our experiments. However, obtaining datasets that meet these requirements is challenging in practice. To overcome this, we employed the SYNLABEL framework~\citep{de2023generating} to generate the realistic soft-labelled datasets for the experiments in Section~\ref{sec:synlabel_exp}. This process is explained in detail in the remainder of this section. Additionally, the methods are tested on the UrinCheck dataset (Section~\ref{sec:urincheck}), for which clean data is unavailable. Consequently, the performance evaluation for UrinCheck was conducted directly on the noisy data.

\subsubsection{Ground Truth Data}\label{sec:ground_truth_data}

To construct synthetic datasets that reflect the realism of real-world data for the experiments in Section~\ref{sec:synlabel_exp}, we follow the steps outlined in the SYNLABEL framework. First, we create a ground truth model $f^G$, which serves as the basis for generating clean datasets. This is done by training a classifier---in this work either LR or RF---on a preprocessed real-world dataset: $D^{RW}=(X^{RW}, y^{RW})$. Thereafter, the original labels, $y^{RW}$, are discarded and replaced by the predictions of the ground truth model on the input data, $y^G = f^G(X^{RW})$. Together, the original feature values, $X^G = X^{RW}$ and the corresponding predicted hard labels, $y^G$, form the Ground Truth dataset $D^G = (X^G, y^G)$.

To generate a soft-labelled dataset, we apply feature hiding (FH): we first determine which features to hide, after which we use the multivariate KDE method to generate marginal distributions for these features. We then sample these distributions 1000 times and replace the original values by the sampled values for the corresponding data points. Next, $f^G$ is used to generate predictions for each of the 1000 resulting versions of the dataset and the multiple predictions for each instance are aggregated into soft labels based on their frequency of occurrence. These labels, $y^{PG}$, together with the features that were not hidden, $X^{PG}$, form a Partial Ground Truth dataset $D^{PG} = (X^{PG}, y^{PG})$. For a more detailed description of the feature hiding method, we refer to the SYNLABEL paper~\citep{de2023generating}.  

In our experiments, for each hard-labelled set $D^G$, we create two soft-labelled sets $D^{PG}$, one with low uncertainty and one with high uncertainty. We quantify this uncertainty by measuring the mean total variation distance ($\overline{TVD}$) between $y^{PG}$ and the one-hot-encoded $y^G$. $\overline{TVD}$ measures the absolute difference two discrete probability distributions $P$ and $Q$, and is defined as:

\begin{equation}\label{eq:tvd}
TVD(P,Q) = \frac{1}{2} ||P - Q||_1.
\end{equation}  

To establish a low and high uncertainty version of a soft-labelled dataset, we first determine which features to hide to obtain a $\overline{TVD}$ value as close to one-third and two-thirds of the way between the $\overline{TVD}$ for hiding one feature and the $\overline{TVD}$ for hiding all but one feature. Features are hidden in order of increasing importance as determined by $f^G$, using either the feature coefficients of the LR model or the feature importances attribute of the RF model.

\subsubsection{Observed Data}

After constructing two types of ground truth datasets, one set with hard labels, $D^G$, and two sets with soft labels, $D^{PG}$, representing uncertainty, noise can be introduced to simulate real-world data. This can be achieved by applying different noise models, further discussed in Section~\ref{sec:noise_types}, to the soft-labelled instances in $D^{PG}$. This results in a noisy, soft-labelled dataset, referred to as an Observed Soft Label dataset $D^{OS} = (X^{O}, y^{OS})$.

To generate a corresponding hard-labelled dataset, a PV or sampling methods can be applied to the soft-labelled instances. This process produces an Observed Hard Label dataset $D^{OH} = (X^{O}, y^{OH})$. These techniques are also used by the HardSingle and HardEns methods to train them directly on the soft-labelled data, bypassing this final step in practice.

\subsection{Introducing Realistic Noise}\label{sec:noise_types}

In real-world applications, data is almost always noisy. While noise can affect both the features and the outcomes, label noise is typically the most detrimental to the predictive performance of trained classifiers~\citep{zhu2004class}. In this study, we investigate how different types of noise affect SLL in particular.

\subsubsection{Label Noise Models}

For hard labels,~\cite{frenay2013classification} distinguishes three different noise models:

\vspace{8pt}
\textbf{Noisy Completely at Random (NCAR)}: in this model, an observed hard label $y^{OH}$ is not equal to its true label $y^{G}$ with probability $p_e$, which represents the strength of the noise. When this happens, a new label is sampled from all other labels with uniform probability. In our study we adapt this model for soft labels by using a transition matrix, where the diagonal entries are $(1-p_e)$ and the off-diagonal entries are $\frac{p_e}{|C| -1}$. This matrix is applied to the original soft label $y^{PG}_{ori}$ to obtain a modified soft label $y^{PG}_{mod}$. Since this results in a deterministic label transformation, we introduce a stochastic component using the following formula: 

\begin{equation}\label{eq:stochastic_application}
y^{OS} = x \cdot y^{PG}_{mod} + (1-x) \cdot y^{PG}_{ori},
\end{equation}

where $x$ is drawn from a normal distribution with $\mu=1.0$ and $\sigma=0.5$. The values in the resulting soft label are truncated to the range [0,1] and $y^{OS}$ is normalised.  

\vspace{8pt}
\textbf{Noisy at Random (NAR)}: for hard labels, the probability that $y^{OH}$ is noisy depends on $y^{G}$. The process we implemented for applying this type of noise is the same as for NCAR , except that the transition matrix is not restricted to have equal off-diagonal values.

\vspace{8pt}
\textbf{Noisy Not at Random (NNAR)}: for hard labels, this type of noise depends on the features $X^{PG}$. This can be modelled in numerous different ways, and will therefore not be considered in this study.

\vspace{8pt}
These noise models assume different dependencies of the noise on the data: no dependence on the data, dependence on only the labels, or dependence on both the features and the labels. These noise models can be applied to both soft and hard labels. For soft labels, however, another type of noise arises naturally that does not translate as well to hard labels: miscalibration.

\subsubsection{Miscalibration}

\begin{figure}[!t]
    \centering
    \includegraphics[width=0.48\textwidth, quiet]{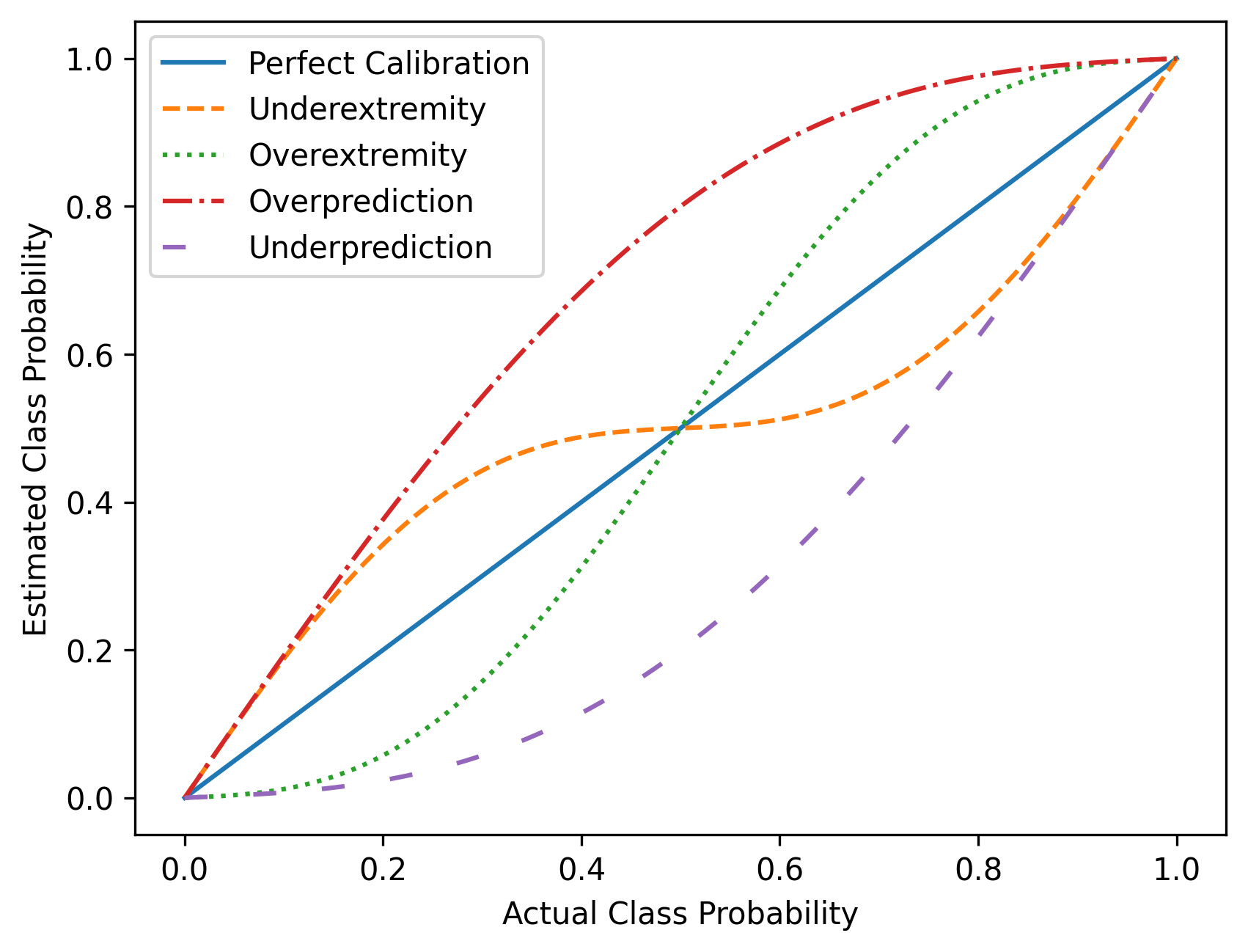}
    \caption{The four miscalibration noise models, defined by Equation~\ref{eq:miscalibration}, for $\beta$ = 0.3.}
    \label{fig:miscalibration_models}
\end{figure}

Human annotators frequently exhibit bias in their estimation of probabilities~\citep{tversky1974judgment, griffin1992weighing}, resulting in miscalibration. Miscalibration is well-studied in the psychology literature, with foundational work by by~\cite{lichtenstein1977calibration}, and expanded upon by  ~\cite{griffin2004perspectives} in which four different miscalibration models are distinguished: 

\begin{itemize}
\item \textbf{Overconfidence}, which includes:
\begin{itemize}
\item \textbf{Overprediction}: Assigning probabilities that are too high.
\item \textbf{Overextremity}: Assigning probabilities that are too extreme (close to 0 or 1).
\end{itemize}
\item \textbf{Underconfidence}, which includes:
\begin{itemize}
\item \textbf{Underprediction}: Assigning probabilities that are too low.
\item \textbf{Underextremity}: Assigning probabilities that are too moderate (close to the midpoint, 0.5).
\end{itemize}
\end{itemize}

Since human annotators are often involved in generating soft labels, miscalibration is highly relevant to this work. Miscalibration models assume that the noise depends only on the probability estimates. As such, miscalibration could be classified as NAR for soft labels. For hard labels, however, this noise depends on additional information, such as a confidence score, which is derived from the characteristics of the instance itself, i.e. dependent on $X$. Therefore, miscalibration is most appropriately placed in the NNAR category.

Figure~\ref{fig:miscalibration_models} illustrates the four types of miscalibration, alongside the line in blue that represent perfectly calibrated probability estimates. The red line denotes the overprediction model, where an annotator assigns probabilities that are consistently higher than the actual class probability. The purple line illustrates underprediction, where probabilities are systematically underestimated. The overextremity model, shown by the green line, is characterized by probability estimates that tend toward the extremes (0 or 1). Finally, the orange line represents the underextremity model, where the probability estimates are too moderate.

We implemented these models using the following formula:

\begin{equation}\label{eq:miscalibration}
\hat{p}(p) = p + \frac{\beta}{\epsilon} \cdot sin(\epsilon \pi p),
\end{equation}

where $\hat{p}$ is the estimated probability, $p$ is the actual probability, $\beta$ represents the strength of the noise, and $\epsilon$ indicates whether there is an extremity. Specifically, $\epsilon = 1$ for over- and underprediction and $\epsilon = 2$ for under- and overextremity. Similarly to the NCAR and NAR models, we apply the miscalibration noise stochastically to the plurality label, via Equation~\ref{eq:stochastic_application}.

In healthcare, miscalibration of probability estimates has been investigated in the context of nursing~\citep{yang2010nurses}, where both over- and underprediction were observed. Furthermore,  ~\cite{xue2017efficient} explored methods to mitigate the effects of noise in soft labels, using both synthetic data and a real-world clinical dataset. In their synthetic data experiments they introduce Gaussian noise into the confidence estimates in their simulated data experiments, corresponding to the NCAR model. In this paper, we investigate how noise models that are specifically tailored to human annotation, the four miscalibration models, affect learning from soft labels, and compare them to the NCAR and NAR noise models.

\section{Realistic Synthetic Data Experiments}\label{sec:synlabel_exp}

In this section, we aim to address two key questions by investigating the performance of different soft label learning (SLL) and hard label learning (HLL) methods on realistic synthetic datasets:

\begin{itemize}
\item Question 1: Can classification model performance be improved by training with soft labels rather than hard labels in the absence of label noise?
\item Question 2: How does noise, and particularly miscalibration, affect the performance of soft label learning?
\end{itemize}

\subsection{Experimental Setup}

To assess the performance of the methods outlined in Section~\ref{sec:algorithms}, we followed the following procedure, making extensive use of the SYNLABEL framework as detailed in Section~\ref{sec:synlabel_data}: for each dataset described in~\ref{sec:datasets}, we first constructed a Ground Truth dataset, $D^G$, with hard labels using a RF as the ground truth model $f^G$. For Question 1, we employed LR as well. From each $D^G$ we constructed two Partial Ground Truth datasets, $D^{PG}$, with soft labels: one with low uncertainty and one with high uncertainty. 

To address Question 1, these datasets $D^{PG}$ were kept noiseless and treated as the observed soft-labelled data, $D^{OS}$. For Question 2, each of the noise models discussed in Section~\ref{sec:noise_types} was applied to $D^{PG}$, with strength $\beta$ increasing from 0.0 to 0.3, resulting in the noisy observed soft-labelled datasets $D^{OS}$. 

Next, we performed 250 random train-test splits, allocating 70$\%$ of the instances as train data and 30$\%$ as test data. All combinations of the methods and base classifiers described in Section~\ref{sec:techniques} were then trained using the train portion of the data and evaluated on the test data as described under Section~\ref{sec:eval}. The methods and code used to run the experiments has been made available on GitHub: \url{https://github.com/sjoerd-de-vries/Soft_Label_Learning}

\subsubsection{Data}\label{sec:datasets}

The 17 datasets used in this study were obtained from the University of California Irvine (UCI)~\citep{Dua:2019} and Knowledge Extraction based on Evolutionary Learning (KEEL)~\citep{Alcala-Fdez2011} repositories. These datasets are listed  in Table~\ref{table:datasets} in Appendix~\ref{appendix:datasets}. We followed the preprocessing procedure that was described in~\cite{de2021reliable}: binary features were binarized, ordinal categorical features were OrdinalEncoded and Standardized, nominal categorical features were OneHotEncoded and numerical features were Standardized. The outcome classes were encoded as integers. The processed datasets are available in the accompanying GitHub repository.

\subsubsection{Evaluation}\label{sec:eval}

Since we generated the datasets used in this section step-by-step, we had access to $D^G$, with clean labels for performance evaluation. Although various metrics can be used, we adopted the area under the curve (AUC) for measuring hard label performance, as it is one of the most widely used classification metrics.

In constructing the soft-labelled datasets $D^{PG}$, we introduced known uncertainty into the labels by hiding specific feature information from the ground truth model. This approach reflects  the uncertainty frequently present in real-world datasets, where not all causal information required for perfect classification is available. Consequently, the most accurate label a model can produce, given the available information, is a discrete probability distribution over the classes, i.e. the soft label $y^{PG}$. To assess how well a classifier predicts this clean soft label we measure $\overline{TVD}$, defined in Equation~\ref{eq:tvd}, between the predicted soft labels and those in $y^{PG}$.

To determine which type of method performs better on noiseless data, answering Question 1, we applied statistical tests to determine if there was a significance performance difference, using $\alpha = 0.05$. Following the recommendations by \cite{garcia2010advanced}, we used the Friedman Aligned-Ranks test for comparing the performance of different methods over multiple datasets. We then applied the Finner post-hoc test to obtain adjusted $p$-values for the comparison of the SoftEns method to HardSingle, SoftSingle and HardEns methods.

\subsection{Noiseless Labels}\label{sec:noiseless_exp}

\begin{figure}[!t]
    \centering
    \includegraphics[width=\textwidth, quiet]{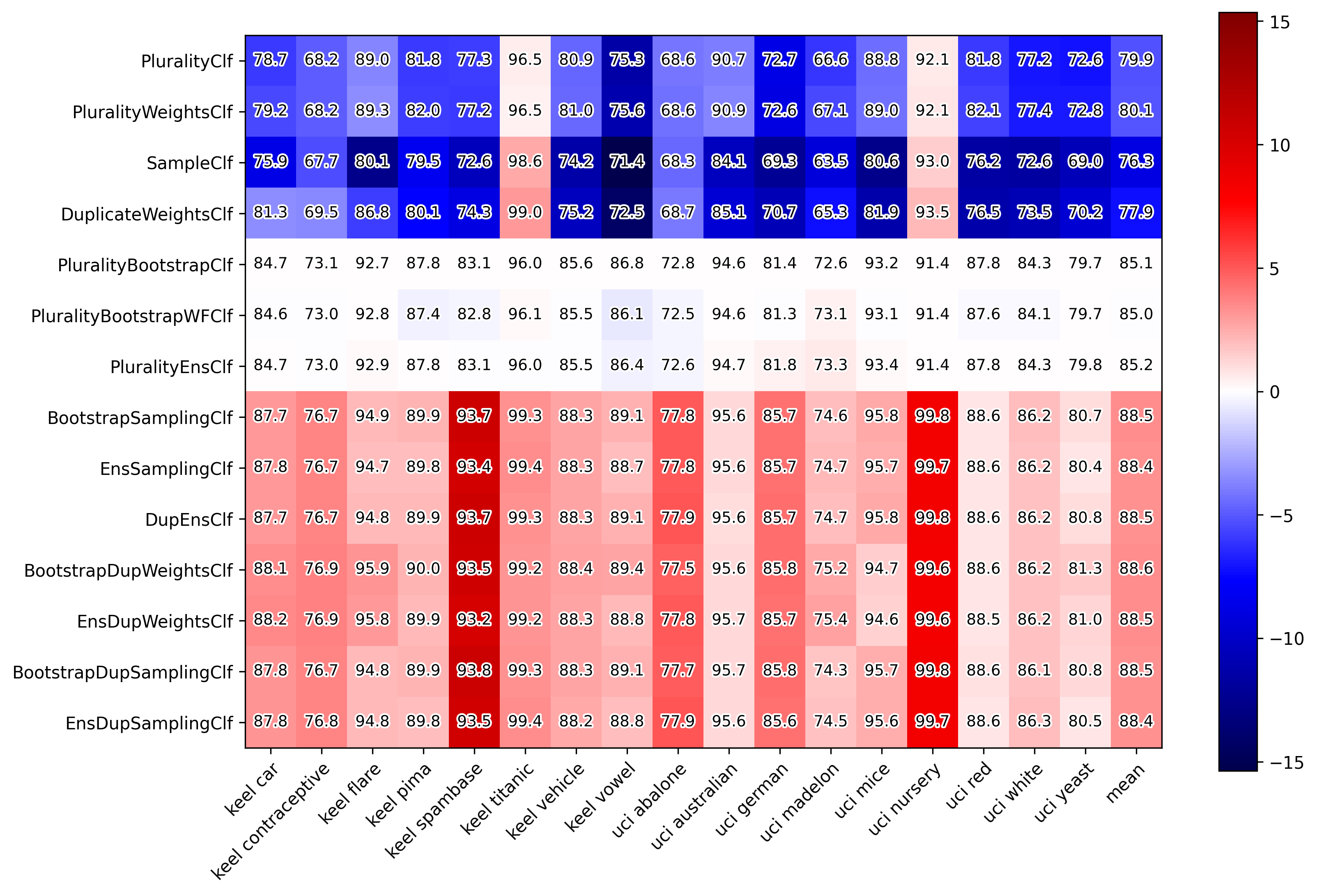}
    \caption{Heat map illustrating the performance of various methods with SGD as base classifier across multiple datasets, along with the their mean performance over all datasets, measured by the AUC on $y^G$. All values were multiplied by $100$ to enhance readability. Red cells indicate higher AUC values, while blue cells represent lower values relative to the AUC of the PluralityBootstrapClf for each dataset.}
    \label{fig:SGD_AUC}
\end{figure}

\begin{figure}[!t]
    \centering
    \includegraphics[width=0.58\textwidth, quiet]{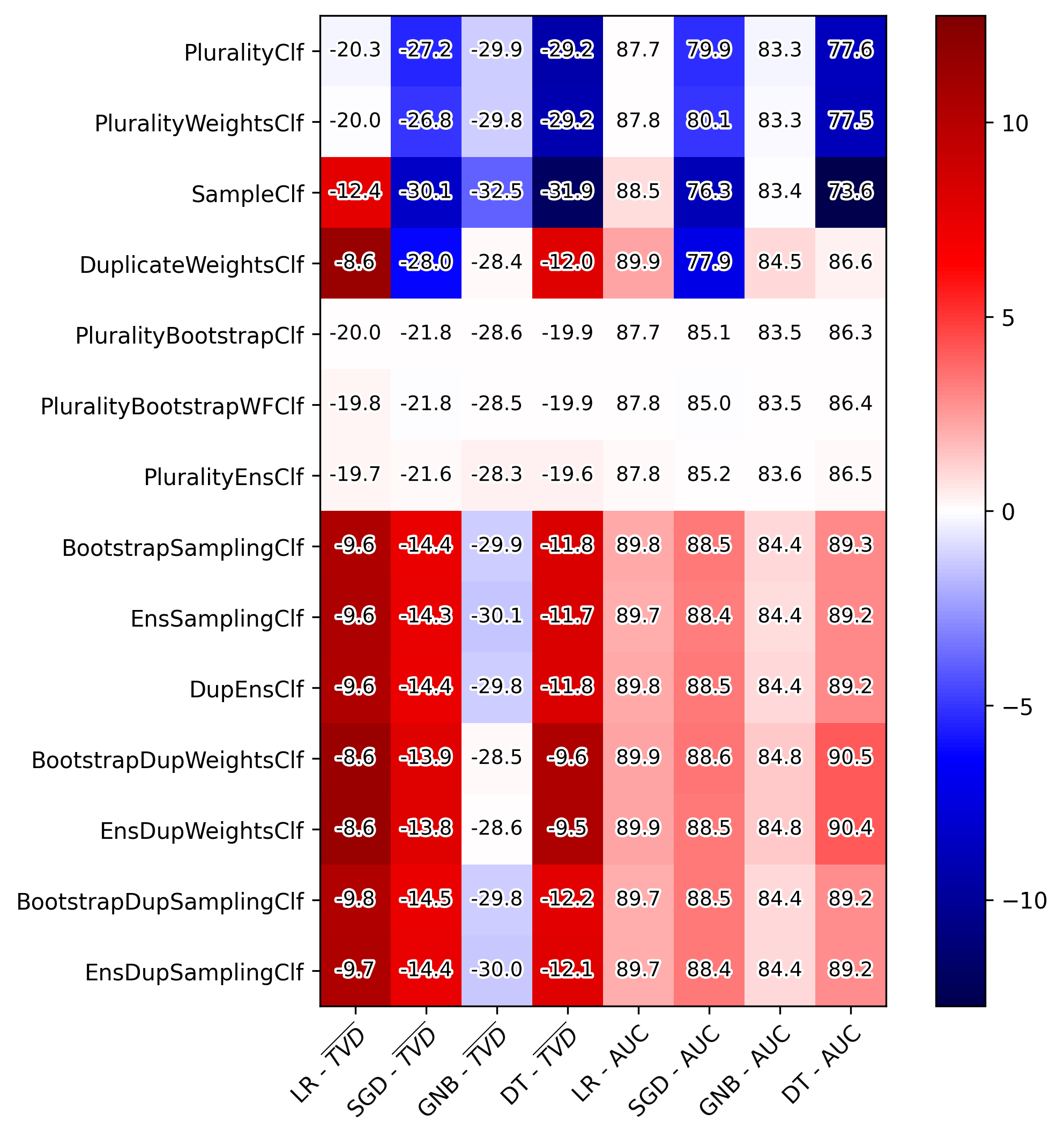}
    \caption{Heat map illustrating the performance of various methods using four base classifier averaged over all datasets. Performance is measured by the AUC on $y^G$ and $\overline{TVD}$ on $y^{PG}$. The $\overline{TVD}$ values were multiplied by $-1$, to allow for easier comparison with the AUC. All values were multiplied by $100$ to enhance readability. Red cells indicate better performance, while blue cells indicate worse performance than PluralityBootstrapClf for each combination of base classifier and metric.}
    \label{fig:stat_table}
\end{figure}

In this section, we aim to answer Question 1: Can classification model performance be improved by training with soft labels rather than hard labels in the absence of label noise?

From the experiments presented in Section~\ref{sec:gaussians}, we concluded that learning from soft labels can be beneficial compared to learning from hard labels, under controlled conditions. However, the results depended on the presence of noise in the soft labels. Here, we investigate whether SLL can result in better performance when dealing with more realistic datasets, starting with noiseless data. In Section~\ref{sec:miscal_exp} we study the effects of noise on these results.

Figure~\ref{fig:SGD_AUC} shows the performance for each of the methods using SGD as base classifier, evaluated over the 17 datasets in terms of AUC relative to $D^G$. The results are averaged over the four datasets resulting from using RF and LR as their $f^G$ with two levels of uncertainty of each soft-labelled dataset. The experiment outcomes are presented in a heat map,  with the color representing the relative performance of a method compared to the PluralityBootstrapClf for a specific dataset. The PluralityBootstrapClf simply applies PV to obtain the most probable class from the soft label and creates an ensemble by employing bagging. We take this as the baseline method, as it is an intuitive method that gives us a realistic indication of the performance that might be expected from a traditional hard label classifier. The methods are ordered such that they match the four categories of learning method, defined in Section~\ref{sec:techniques}: HardSingle, SoftSingle, HardEns, SoftEns. 

For this specific combination of SGD as base classifier and AUC as evaluation metric, we clearly observe that the ensemble methods outperform the single classifier methods. Furthermore, the SoftEns methods have the best overall performance, surpassing their HardEns counterparts. The performance of the SoftSingle methods is worse than the HardSingle methods, on average, although for some datasets the reverse is true. The DuplicateWeightsClf outperforms the SampleClf on every dataset, which is unsurprising as that latter simply samples a single hard label according to a soft label. This classifier would hardly be used in practice, as taking the most probable class as the PluralityClf does, is more intuitive.

Results for the other seven combinations of a base classifier (LR, SGD, GNB, DT) and metric (AUC, $\overline{TVD}$), are presented in Figures~\ref{fig:q1_LR_AUC}-\ref{fig:q1_DT_TVD} in Appendix~\ref{appendix:q1_heatmaps}. In general, the patterns are similar to those for the SGD-AUC setting, although the SoftSingle methods, and especially the DuplicateWeightsClf, perform considerably better with LR, GNB and DT as base classifier. Additionally, the impact of ensemble learning is smaller for LR and GNB, leading to similar performance between the HardSingle and HardEns methods, as well as between the SoftSingle and SoftEns methods.  In terms of the $\overline{TVD}$ measured on $y^{PG}$, we observe broadly similar patterns to those seen for the AUC measured on $y^G$. 

To conclusively answer the first question, we further average the scores over all datasets to obtain a score for each method and for the combination of a base classifier with each of the two metrics: AUC on $D^G$ as before and $\overline{TVD}$ on $D^{PG}$. The results are presented in Figure~\ref{fig:stat_table}. We took the negative of $\overline{TVD}$, so higher scores indicate better performance, for easier visual comparison with the AUC.

We observe that the SLL methods and especially the SoftEns methods outperform the HLL methods with the LR, SGD and DT base classifiers. The results for the GNB base classifier are inconclusive, where the SoftEns methods outperform the HardEns methods when we look at the AUC, while the HardEns methods perform better when measuring $\overline{TVD}$. From this aggregated table, we observe that the DuplicateWeightsClf outperforms the HardSingle and HardEns methods for LR and DT, while having comparable performance with GNB and worse performance with the SGD base classifier. The SampleClf again has worse performance than the DuplicateWeightsClf, but manages to outperform the HardEns methods with LR. 

To ensure the statistically validity of these findings, we applied the Friedman Aligned-Ranks test to the best performing method in each of the four categories: PluralityClf, DuplicateWeightsClf, PluralityEnsClf and BootstrapDupWeightsClf. This analysis was conducted using all 17 datasets, both ground truth models (LR and RF) and both uncertainty levels, for a total of 68 comparisons. The results are shown in Table~\ref{table:stat_tests}. 

We observe that for the LR, SGD and DT base classifiers, the hypothesis that all methods perform equally well is rejected under significance level of $\alpha=0.05$. For GNB, however, this is not the case. In terms of the ranks, we observe that the SoftEns method has the lowest rank for $\overline{TVD}$, where lower is better, and the highest for AUC, where higher is better, for all of the base classifiers, except for LR where the SoftSingle classifier obtains a slightly lower rank (71.79 versus 72.56). Additionally, the $p$-value of the direct comparison via the Finner test is below 0.05, except for with GNB as base classifier and the combination of SoftSingle with LR. 

In conclusion, learning from soft labels under noiseless conditions can lead to performance improvements, as the best SoftEns method significantly outperforms both the best HardSingle and HardEns methods for three out of four base classifiers, while achieving equal performance with the fourth.

Additionally, we tested methods that incorporated thresholding. These are shown in Appendix~\ref{appendix:threshold}, where the number behind the name of the classifier indicates the percentage of the instances that was discarded, starting from the most uncertain instances, i.e. the percentage threshold. We show the performance over all classifiers and both AUC and $\overline{TVD}$ in Figure~\ref{fig:stat_table_threshold}. We observe that for these average results, none of the thresholding methods has better performance than PluralityBootstrapClf for any of the settings, with a tendency toward better results when a lower number of data points is excluded. In the following we therefore exclude the methods using thresholding.

\begin{table*}[ht]
\centering
\caption{The ranks and $p$-values from statistical tests comparing the performance of different methods across multiple datasets. The ranks are based on the Friedman Aligned-Ranks test. Lower ranks indicate better performance for $\overline{TVD}$ while higher ranks indicate better performance for AUC. Adjusted $p$-values from the Finner test are provided to assess the statistical significance of the differences between methods, with values below 0.05 indicating significant differences.}
\label{table:stat_tests}
\begin{adjustbox}{max width=\textwidth}

\begin{tabular}{llrrrrrrrr}
\hline
 &  & $p$-value & \multicolumn{4}{r}{Rank} & \multicolumn{3}{r}{Adjusted $p$-value: SoftEns} \\
 &  & Friedman & HardSingle & SoftSingle & HardEns & SoftEns & vs HardSingle & vs SoftSingle & vs HardEns \\
 \hline
\multirow[c]{4}{*}{$\overline{TVD}$} & LR & 0e0 & 203.93 & 71.79 & 197.72 & 72.56 & 0e0 & 9.55e-01 & 0e0 \\
 & SGD & 0e0 & 189.33 & 201.28 & 111.49 & 43.90 & 0e0 & 0e0 & 5.43e-07 \\
 & GNB & 5.58e-02 & 161.69 & 133.09 & 128.22 & 123.00 & 1.23e-02 & 5.97e-01 & 6.99e-01 \\
 & DT & 0e0 & 228.75 & 90.84 & 161.44 & 64.97 & 0e0 & 5.52e-02 & 1.29e-12 \\
 \hline
\multirow[c]{4}{*}{AUC} & LR & 2.26e-14 & 88.21 & 180.42 & 95.23 & 182.14 & 1.00e-11 & 8.99e-01 & 1.76e-10 \\
 & SGD & 0e0 & 89.26 & 64.84 & 172.85 & 219.05 & 0e0 & 0e0 & 6.15e-04 \\
 & GNB & 3.43e-02 & 112.27 & 139.82 & 136.33 & 157.57 & 2.35e-03 & 1.88e-01 & 1.68e-01 \\
 & DT & 0e0 & 39.96 & 138.70 & 149.57 & 217.77 & 0e0 & 6.89e-09 & 4.29e-07 \\
 \hline
\end{tabular}

\end{adjustbox}
\end{table*}

\subsection{Miscalibration}\label{sec:miscal_exp}

In Section~\ref{sec:gaussian_noise}, we observed that adding noise negatively impacts the performance of SLL. While the previous experiments demonstrated the potential benefits of SLL in noiseless conditions, in this section we address the question: How does noise, and particularly miscalibration, affect the performance of soft label learning?

Our earlier experiments revealed that method performance varied more with different levels of uncertainty in the soft labels using the same ground truth model than across different ground truth models. To reduce the computational cost, we therefore chose to use only RF as the $f^G$ for these experiments, while varying the level of uncertainty in the soft labels in $D^{PG}$.

We introduced six types of noise into $D^{PG}$ to obtain $y^{OS}$ and $D^{OS}$, as described in Section~\ref{sec:noise_types}: NCAR, NAR, overprediction, underprediction, overextremity and underextremity. The strength of the noise, $\beta$, was varied between 0.0 (level 0) and 0.3 (level 6). Importantly, $\beta$ scales differently for the different noise models, which makes a direct comparison between the impact of different noise types impossible. Unlike in the experiment with Gaussian distributions in Section~\ref{sec:gaussians}, the noise applied here affects HLL as well.

\begin{figure}[p]
    \centering
    \includegraphics[width=\textwidth, quiet]{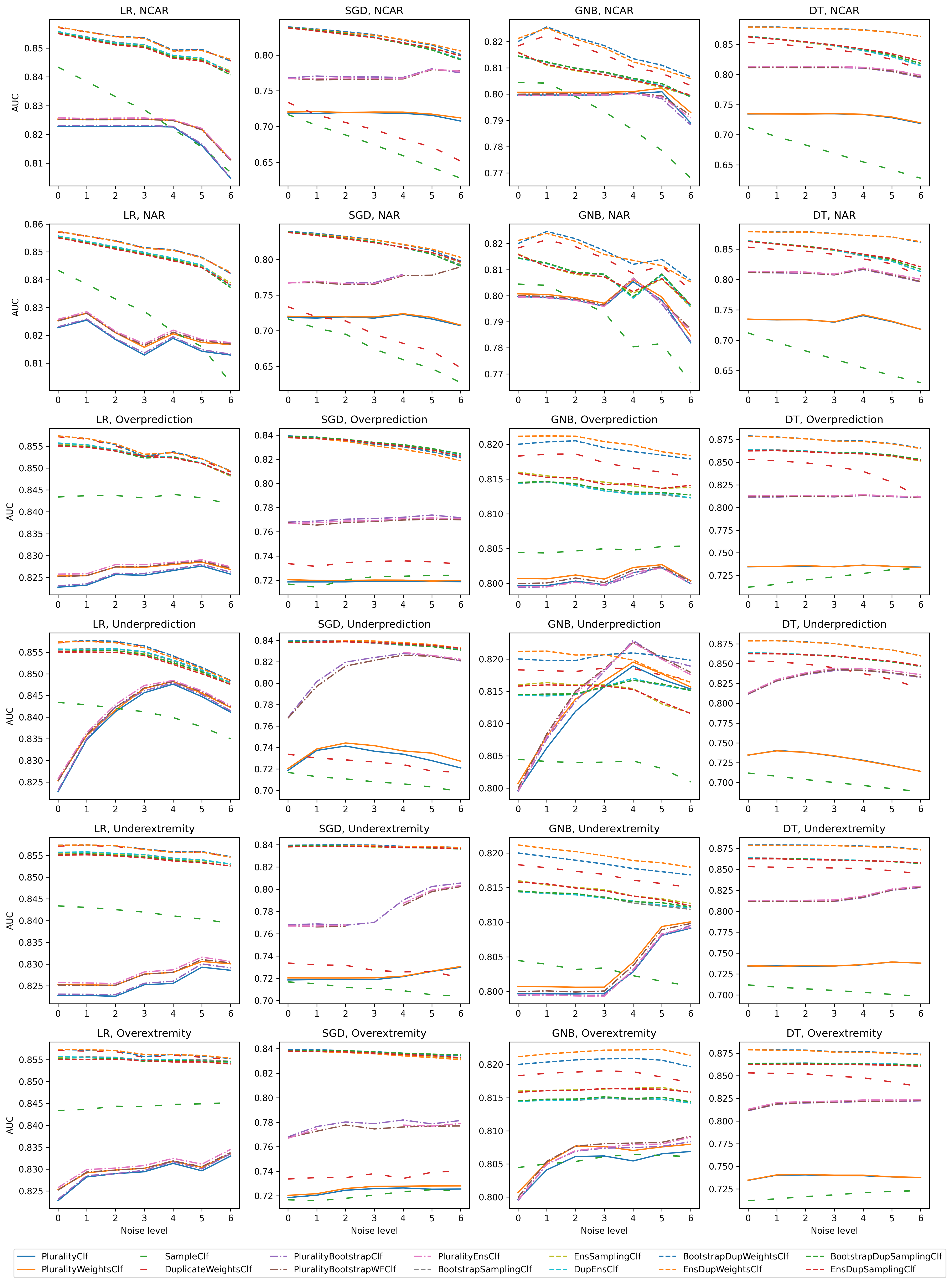}
    \caption{The effect of six different noise types on method performance with four different base classifiers, measured by the AUC on the ground truth test data, across multiple noise levels. Noise types include NCAR, NAR, overprediction, underprediction, underextremity and overextremity, with noise levels ranging from level 0 (noiseless) to level 6 (noise strength 0.3). LR, SGD, GNB and DT were used as base classifiers. RF served as the ground truth model, with the soft labels generated at the high uncertainty level.}
    \label{fig:q2_rf2_auc}
\end{figure}

The results for a $D^G$ constructed with RF under the high uncertainty setting are shown in Figure~\ref{fig:q2_rf2_auc}, for each of the four base classifiers used throughout this work and each of the six noise types. The figure shows the AUC measured on the test data in $D^G$ on the $y$-axis and the noise levels are shown on the $x$-axis. The results for a $D^{PG}$ containing lower uncertainty, as well as for performance measured in $\overline{TVD}$ with respect to $y^{PG}$, are shown in Figure~\ref{fig:q2_rf1_auc}-\ref{fig:q2_rf2_tvd} in Appendix~\ref{appendix:q2}. In the following analysis we mainly focus on Figure~\ref{fig:q2_rf2_auc}, although we will reference the other figures when they exhibit different patterns. In Figures~\ref{fig:q2_rf1_auc_base}-~\ref{fig:q2_rf2_tvd_base} in Appendix~\ref{appendix:q2_base} the difference in performance between the classifier trained on the noiseless data and the classifier trained on the noisy data is shown, to further highlight the specific impact of the noise.

\subsubsection{Baseline Performance}

We first analyse method performance in the baseline setting, without noise (level 0). This performance remains constant for each classifier across the different noise type plots.

In the high uncertainty setting, the SoftEns methods consistently achieve the highest baseline AUC across the base classifiers. In contrast, the HardSingle methods rank near the bottom. The two SoftSingle methods differ substantially in performance: DuplicateWeightsClf performs poorly with the SGD base classifier, as we observed in Section~\ref{sec:noiseless_exp}, but achieves an AUC that is competitive with the SoftEns method otherwise. Meanwhile, SampleClf ranks among the worst performing methods, showing some promise only when combined with LR. The HardEns methods outperform the HardSingle methods when paired with SGD and DT, and perform comparably for LR and GNB, but still fall short of the AUC achieved by the SoftEns methods. 

In the low uncertainty setting (Figure~\ref{fig:q2_rf1_auc}), we largely observe the same order in the method performances. However, overall performance improves significantly, with the HardEns methods showing particularly strong gains compared to the high uncertainty setting, approaching the performance of the SoftEns methods. The substantial drop in performance of the HardSingle and HardEns methods for the higher uncertainty setting can be attributed to the increased in uncertainty in the soft labels, which leads to a greater number of incorrect hard labels. While the SLL methods explicitly take this uncertainty into account, the HLL methods  are trained on a higher proportion of incorrect labels without utilizing knowledge of the underlying uncertainty, having a larger negative impact on their their performance.

When examining performance as measured by $\overline{TVD}$ on $y^{PG}$ in Figures~\ref{fig:q2_rf1_tvd} and ~\ref{fig:q2_rf2_tvd}, the ranking of the methods is largely preserved. However there are notable differences for the GNB base classifier, where the HardEns methods see a relative improvement in performance.

\subsubsection{Impact of Random Noise}
Next, we investigate the effect of different noise types on method performance. The performance change relative to the noiseless setting (level 0) is shown separately for both metrics and uncertainty settings in Figures~\ref{fig:q2_rf1_auc_base}-\ref{fig:q2_rf2_tvd_base} in Appendix~\ref{appendix:q2_base}.

For the NCAR setting, we largely observe similar performance patterns across the different base classifiers. The SampleClf is most affected by this type of noise. While it already exhibited the worst performance for SGD and DT on the noiseless data, a further sharp performance decrease is caused by the NCAR noise. The SampleClf had better baseline performance for LR and GNB, but due to the large effect the NCAR noise has, it is among the worst performing methods at large noise levels for these base classifiers as well. 

The DuplicateWeightsClf method follows the same pattern when combined with the SGD classifier, while for the other three base classifiers it performs much better, in line with the SoftEns methods. When we look at $\overline{TVD}$, however, the performance of the SoftEns and DuplicateWeight methods is impacted by the NCAR noise as much as the SampleClf. 

Furthermore, across metrics the SoftEns and Soft methods are impacted by this noise more than the HardEns and HardSingle methods. Nevertheless, for the high uncertainty setting SoftEns AUC remains better than that of the HardEns methods, as does the DuplicateWeightsClf except for with SGD. For the low uncertainty settings HardEns performs similar or slightly better than the SLL methods at high noise levels. In terms of $\overline{TVD}$, for low uncertainty and high noise levels the HLL methods outperform the SoftEns methods as well. For high uncertainty the performance differs per base classifier, with the HardEns method performing better overall. 

For the NAR setting, the results and conclusions are very similar to those of the NCAR setting. 

\subsubsection{Impact of Miscalibration}
Next, we examine the types of noise that simulate a miscalibrated annotator, as defined in Section~\ref{sec:noise_types}.

The SoftEns methods are affected by each type of miscalibration noise to varying degrees, but generally behave as expected across both metrics and uncertainty levels: as noise increases, performance declines. An interesting exception to this occurs in the low uncertainty setting, where $\overline{TVD}$ initially improves for the overprediction and overextremity noise models.
 
For the SoftSingle methods, the DuplicateWeightsClf generally follows the same trends as the SoftEns methods for LR and GNB, while the noise has a larger impact for the DT and SGD classifiers. Notably, for SGD the performance improves with increasing noise for the overprediction and overextremity models. SampleClf further experiences improvement across metrics and uncertainty settings for the overextremity and overprediction models.
 
Interestingly, the HardSingle and HardEns methods frequently improve their AUC at the lower noise levels, after which performance seems to stagnate or decline, for both the high and low uncertainty settings. When considering $\overline{TVD}$ under high uncertainty, performance frequently improves or is remains stable at lower noise levels, whereas for lower uncertainty performance decreases for the overprediction and overextremity. Broadly, the reaction of the HLL methods to the miscalibration noise models at the lower noise levels is the opposite of that of the SLL methods. When the performance of HLL increases, SLL performance stagnates or decreases and vice versa. At the higher noise levels, however, nearly all methods start to become negatively affected by the noise or have their performance stagnate. 

Despite these varied reactions to the different noise types, especially at the lower noise levels, the SoftEns methods achieve the best performance for all miscalibration noise settings, in terms of AUC. This remains true when $\overline{TVD}$ is considered for the high uncertainty setting, whereas for the low uncertainty setting the performance of the SoftEns and HardEns methods is more balanced.

To summarise, for noiseless data or data affected by miscalibration, the SoftEns methods clearly outperform the HardEns methods. In case of random noise, the HardEns methods achieve higher performance when the data contains low base levels of uncertainty, whereas the SoftEns methods perform better if the data are more uncertain before the introduction of additional noise.

\section{Real-World Data Experiment}\label{sec:urincheck}

Having demonstrated the performance of the different SLL and HLL methods on synthetic, real-world inspired data, we now apply them to an actual real-world dataset. Unlike in the previous experiments, there are no clean ground truth or partial ground truth labels available for most real-world datasets, presenting challenges for method evaluation.

It is not immediately clear which labels should be used for measuring performance. When confidence scores or soft labels are available, it is common practice to convert them to hard labels by assigning the class with the highest probability (plurality vote). However, this approach overlooks the uncertainty inherent in the soft labels. A data point may be assigned a hard label with either 100$\%$ certainty or the assigned class may have only marginally higher probability than the others. Arguably, sampling the labels according to the soft label probabilities is a more representative approach. Alternatively, the soft labels themselves can be used for evaluation, although this prevents the use of common classification metrics such as AUC and accuracy. Regardless of the label set that is chosen, a real-world dataset may contain unspecified noise, meaning that any hard label in the evaluation set could be incorrect, or the probabilities in the soft labels might be inaccurate.

For this experiment involving real-world data, we use the UrinCheck dataset, as introduced by ~\cite{de2022semi}, as an example of a dataset for which confidence scores, and by extension soft labels, are available. This dataset was collected with the goal of predicting urinary tract infection (UTI) based on patient demographics (age, sex) and  laboratory values. In clinical practice, urine culture results alone are insufficient to diagnose a UTI, as the presence of symptoms is also required. As a result, labels could not be assigned automatically to the data as this required interpreting the written information describing these symptoms contained in the electronic health records (EHR). Additionally, the presence of a UTI was not recorded in a structured format in the EHR, necessitating the retrospective creation of a labelled dataset through expert annotation by medical professionals. Each expert provided a hard label along with a confidence score $c$ in the range from 5 to 10. If $c$ was 6 or below, a discussion with another expert followed until consensus was reached, ensuring that all final labels had a confidence score of at least 6. This annotation process resulted in a dataset of 906 labelled cultures, from 810 patients, each with both a hard label and a confidence score. Since some culture data were missing an important predictor, the urinalysis, we focus on the subset of 717 labelled cultures for which the urinalysis data is available. Given that this is a binary classification problem, the confidence scores can be easily converted into soft labels. Specifically, for hard label ``no UTI" the soft label becomes $[c', 1-c']$, and for hard label ``UTI" it becomes $[c', 1-c']$, with $c' = \frac{c}{10}$.

\begin{figure}[!t]
    \centering
    \includegraphics[width=\textwidth, quiet]{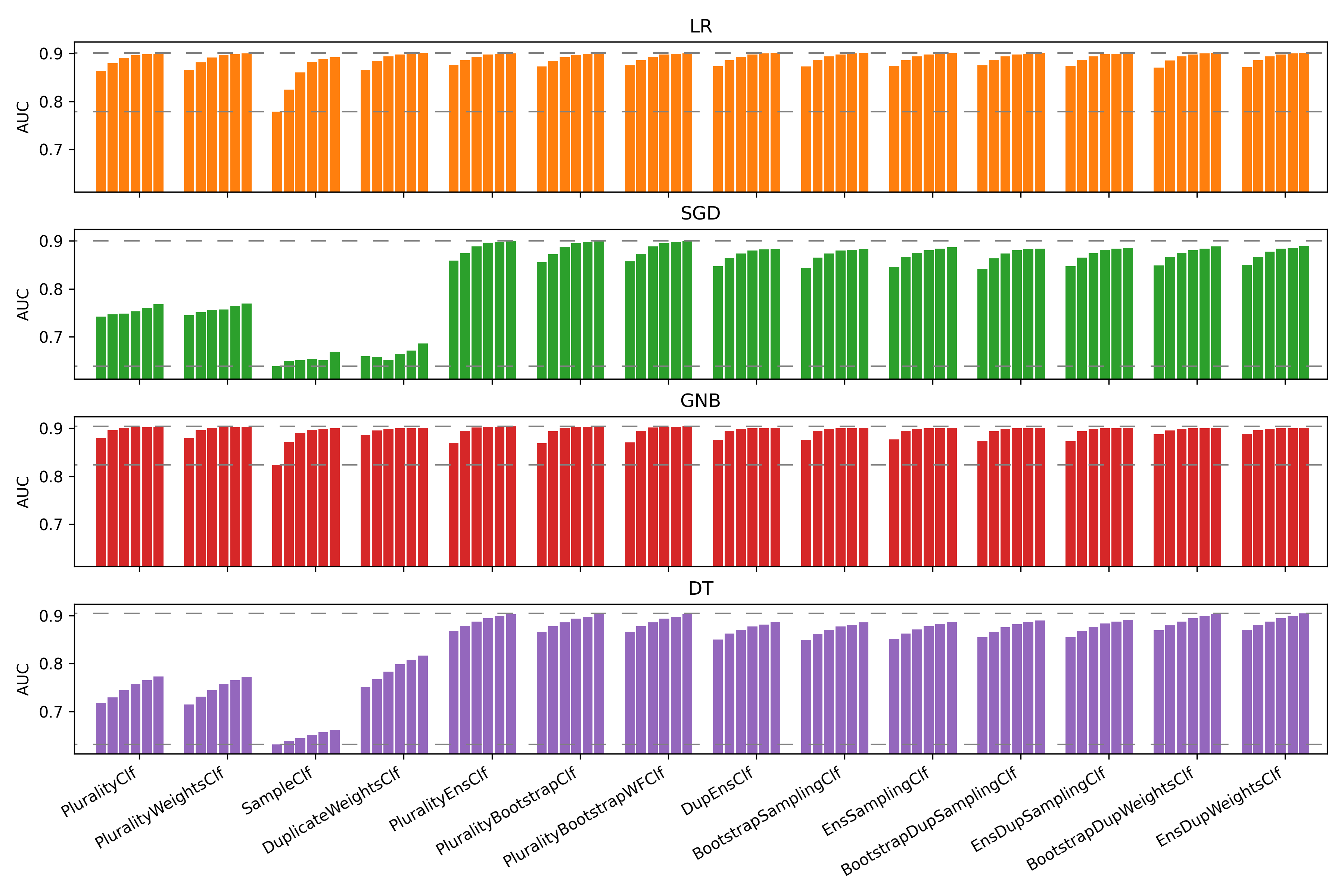}
    \caption{Method performance on the UrinCheck dataset, measured by AUC. The labels used for the test set are the most probable labels from the soft-labelled data. For each method the individual bars represent the fraction of the total data that was used as training data in that experiment, increasing from left to right: $\{0.05,0.1,0.2,0.4,0.6,0.8\}$}
    \label{fig:rw_auc_pv}
\end{figure}

To investigate the benefits of using soft-label methods for training a classifier, we applied the same methods and base classifiers as in Section~\ref{sec:miscal_exp} to the UrinCheck dataset. We evaluated their performance by measuring the AUC on the hard labels and the $\overline{TVD}$ on the soft labels. Two sets of hard labels were used for the evaluation in terms of AUC: either the original hard label assigned by the experts was used, which corresponded to the class with the highest probability in the soft label, or the label was sampled according to the soft label probabilities. We varied the amount of labelled data provided to the classifiers to asses the influence of data availability. The portion of the data used in training was ranged from 5$\%$ to 80$\%$, with the remainder of the data used as the test set. Each experiment was repeated 1000 times.

\begin{figure}[!t]
    \centering
    \includegraphics[width=\textwidth, quiet]{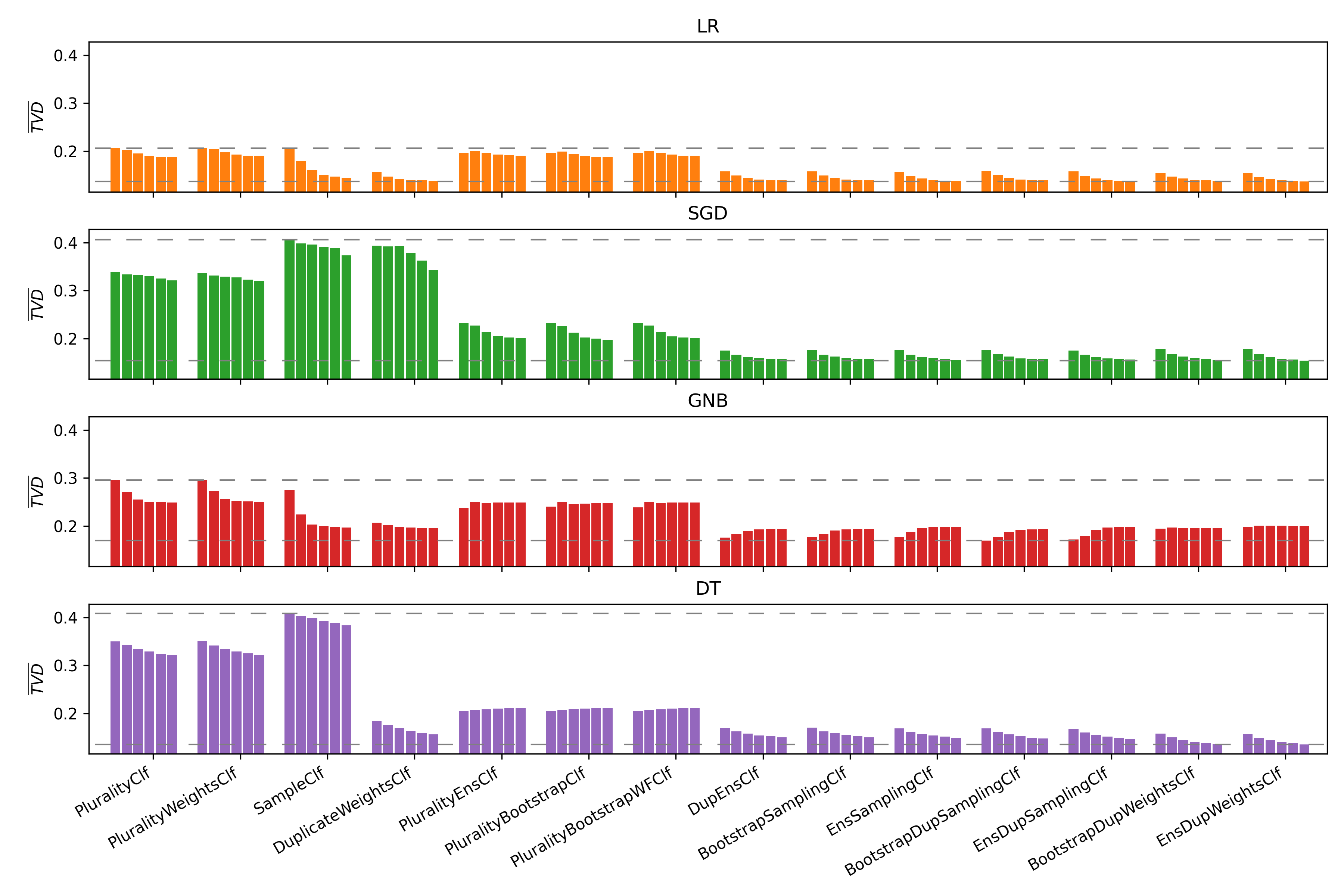}
    \caption{Method performance on the UrinCheck dataset, measured by $\overline{TVD}$. The labels used for the test set are the original soft labels. For each method the individual bars represent the fraction of the total data that was used as training data in that experiment, increasing from left to right: $\{0.05,0.1,0.2,0.4,0.6,0.8\}$}
    \label{fig:rw_tvd_soft}
\end{figure}

Figure~\ref{fig:rw_auc_pv} shows the performance of the different methods, measured by the AUC on the original hard labels. 

The performance patterns for most methods are nearly interchangeable between the SGD and DT base classifiers, with the exception of the DuplicateWeightsClf, which performs considerably worse with SGD. Furthermore, the SampleClf performs particularly poorly with both SGD and DT compared to its performance with LR and GNB. These trends are consistent with the results observed on synthetic data. We observe that the ensemble learning methods perform much better than the single classifier methods. Furthermore, the HardEns methods slightly outperform the SoftEns methods for SGD and are comparable in performance to the Bootstrap-/EnsDupWeightsClassifier for DT, while outperforming the other SoftEns methods. Among the single classifier methods, PluralityClf and PluralityWeightsClf clearly outperform the DuplicateWeightsClassifier for SGD, while for DT the latter has better performance. 

The performance patterns for the LR and GNB base classifiers resemble each other as well. The single classifier methods perform much better with LR and GNB compared to SGD and DT, obtaining similar AUC as the ensemble methods, with the exception of the SamplingClf, especially for the lower training fractions. This can likely be attributed to the LR and GNB classifiers having lower variance than the SGD (SVM) and especially DT classifiers, which reduces the potential benefits of ensemble learning. 

As expected, the performance for each of the classifiers and learning methods increases as more training data is made available to them. There is no clear indication that the soft-labelled data classifiers perform better under the low data settings, which might be due to the fact that good class separation is already possible using very few data points for this particular problem. This is supported by the relatively small performance difference between training on 5$\%$ or 80$\%$ of the data across almost all base classifier and method combinations. 

In summary, when evaluated using AUC with the noisy hard labels from the UrinCheck dataset as target, the best SoftEns methods perform equally to the best HardEns methods for three out of the four base classifiers. The best overall AUC for a a combination of base classifier and method is achieved by both SoftEns methods and HardEns methods with DT as a base model, indicating that the best-case performance is equivalent.

In Figure~\ref{fig:rw_tvd_soft}, the performance on the soft labels is presented as measured by $\overline{TVD}$, where lower values indicate better performance. This provides a stark contrast to the previous results: all SLL methods clearly outperform HLL, except for the DuplicateWeightsClassifiers with SGD and the SampleClf with both SGD and DT, which we had already observed to be poor combinations in the previous experiments.

Thus, while using SLL methods does not necessarily lead to better class predictions when evaluated against the suboptimal, noisy hard labels, the performance of especially BootstrapDupWeightsClf and EnsDupWeightsClf is comparable. More importantly, the predicted probabilities are significantly better calibrated with respect to the original soft labels (expert confidences) for these data.

In Figure~\ref{fig:rw_auc_samp} in Appendix~\ref{appendix:rwd}, method performance is shown on a test set for which the hard labels are not the most probable class, but are instead sampled according to the class probabilities for each data point. While in practice this approach for evaluation is  generally inadvisable, as the results for each sampled label set would be highly variable, and assigning less probable class labels is counter-intuitive, the resulting sets are a reflection of what the true labels might look like.

As expected, method performance on this set is significantly worse compared to the regular hard labels. Interestingly, the overall performance pattern remains largely the same, with the notable exception that the Bootstrap-/EnsDupWeightsClf now emerge as the best performing methods for DT and LR, outperforming the hard label ensemble methods.   

\section{Conclusion}

This study explored the value of incorporating information about the uncertainty of the outcome---soft label learning (SLL)---in classification models compared to the standard approach of hard label learning (HLL). To this end, we evaluated a number of different wrapper methods for HLL and SLL, using the same base classifiers to ensure fair comparisons. Our experiments considered simulated data, realistic synthetic data and a real-world dataset, with a particular focus on assessing the impact of different types of label noise, including four types of miscalibration.

The simulated data experiment showed that using soft labels greatly improved the accuracy of estimated model parameters compared to using hard labels, especially in cases of limited data and class imbalance. This suggested there is potential value in using soft label information to enhance classification models. Notably, even when noise was introduced exclusively into the soft labels, their advantage persisted for small sample sizes.

In experiments with realistic synthetic datasets, where both clean hard and soft labels were available for evaluation, SLL was consistently better than HLL in the noiseless setting, significantly outperforming them for three out of four base classifiers. Similar performance was observed for under noisy conditions: the ensemble SLL approaches generally achieved better performance than the HLL approaches in predicting clean hard labels across all noise levels. For predicting the soft labels, this trend held true in most settings, although the HLL ensemble approaches occasionally did better. Further analysis revealed notable differences in how noise affected the different methods, particularly between between the traditional noise models (Noisy Completely At Random and Noisy At Random) and the newly introduced miscalibration noise types. Overall, the SLL approaches were affected slightly more by noise, but nevertheless had better performance than the HLL methods even for the higher noise levels, largely due to their superior baseline performance on the noiseless data.

For the real-world data, only the noisy labels that were obtained during the labelling process were available. While this makes for a less than optimal evaluation setting, it reflects the scenario frequently encountered in practice. On this dataset, we found that the SLL and HLL methods performed similarly for predicting hard labels, whereas the soft label methods had much better performance for predicting the soft label, demonstrating better calibration. Notably, no clear differences in performance were observed for different sample sizes.

In conclusion, the SLL methods nearly always outperformed the HLL methods, or at least had comparable performance across all three types of experiments. This underscores the need for further research into SLL methods. Additionally, this work suggests that investing in obtaining uncertainty information during an annotation process, which is often relatively inexpensive, can lead to significantly improved classification models.

\clearpage

\backmatter

\begin{appendices}

\section{Datasets} \label{appendix:datasets}

\begin{table*}[ht]
\centering
\caption{Dataset characteristics. This table summarizes the 17 datasets sourced from the UCI and KEEL repositories that were used in the experiments of Section~\ref{sec:synlabel_exp}. The dataset descriptions reflect their characteristics after preprocessing.}
\label{table:datasets}
\makebox[0.8\textwidth][c]{
\begin{tabular}{ l c c c c}
    \hline
    Data set & Size & Variables & Classes & Class balance $(\%)$ \\ \hline
    Abalone\footnotemark[1] & 4177  & 10 & 3 & 35/34/32\\	
	Australian Credit & 690 & 39 & 2 & 56/44 \\
    Car Evaluation & 1728  & 6 & 4 & 70/22/4/4 \\
    Contraceptive Method Choice & 1473  & 9 & 3 & 43/35/23 \\
    German & 1000  & 52 & 2 & 70/30\\
    Madelon & 2600  & 500 & 2 & 50/50\\
	Mice Protein Expression & 1080  & 77 & 8 & 14/14/13/13/13/13/13/10\\	
	Nursery & 12958  & 10 & 4 & 33/33/31/3\\
	Pima Indian Diabetes & 768  & 8 & 2 & 65/35 \\ 
	Red wine\footnotemark[2] & 1599  & 11 & 2 & 53/47\\ 	
    Solar Flare & 1066  & 37 & 6 & 31/22/20/14/9/4 \\
    Spambase & 4597  & 57 & 2 & 61/39\\
    Titanic & 2201 & 6 & 2 & 68/32\\
    Vehicle Silhouettes & 846  & 18 & 4 & 26/26/25/24 \\
    Vowel & 990  & 13 & 11 & 9/9/9/9/9/9/9/9/9/9/9\\
	White wine\footnotemark[2] & 4898 & 11 & 2 & 67/33\\ 	
	Yeast & 1484 & 8 & 10 & 31/29/16/11/3/3/2/2/1/0 \\
	\hline 
\end{tabular}}
\end{table*}

\footnotetext[1]{Transformed into a classification problem with 3 ring (age) outcome classes: $<$9, 9$-$10, $>$10, as suggested in~\cite{Macia2014}.}
\footnotetext[2]{Transformed into the binary problem of grading wines with outcome either $<$6 or $\geq$6. Datasets from~\cite{cortez2009modeling}.}

\clearpage

\section{Base Classifiers}\label{appendix:classifiers}

\begin{table*}[h]
\renewcommand{\arraystretch}{1.1} 
\caption{Base classifier parameter values. GNB: Gaussian Naive Bayes, SGD: Stochastic Gradient Descent (equal to a quadratically smoothed Support Vector Machine (SVM) with $\gamma = 2$), DT: Decision Tree, LR: Logistic Regression. Parameters not listed here were kept at their default values as implemented by the scikit-learn package~\citep{scikit-learn}.}
\vspace*{5mm}
\label{tab:classifiers}
\centering
\begin{tabular}{l c}
    \hline
    Classifier (parameter) & Value  \\ \hline
    \textbf{GNB} & \\
    ~~all & default \\ 
    \textbf{SGD} & \\
    ~~loss & modified huber \\
    ~~early stopping & True for large sets \\
    ~~penalty & L2 \\
    ~~alpha & 0.0001 \\
    \textbf{DT} & \\
    ~~criterion & gini \\
    ~~splitter & best \\
    ~~max depth & None \\
    ~~features & $\sqrt{n_{features}}$ \\
    \textbf{LR} & \\
    ~~penalty & L2 \\
    ~~C & 1 \\
    \hline
  \end{tabular}
\end{table*}

\clearpage

\section{Question 1: Heat Maps} \label{appendix:q1_heatmaps}

Shared description for all heat maps in Appendix C: Heat map illustrating the performance of various methods with the mentioned base classifier across multiple datasets, along with the their mean performance over all datasets, measured by either AUC on $y^G$ or $\overline{TVD}$ on $y^{PG}$. All values were multiplied by $100$ to enhance readability. Red cells indicate higher AUC values, while blue cells represent lower values relative to the performance of the PluralityBootstrapClf for each dataset.

\begin{figure}[h]
    \centering
    \includegraphics[width=\textwidth, quiet]{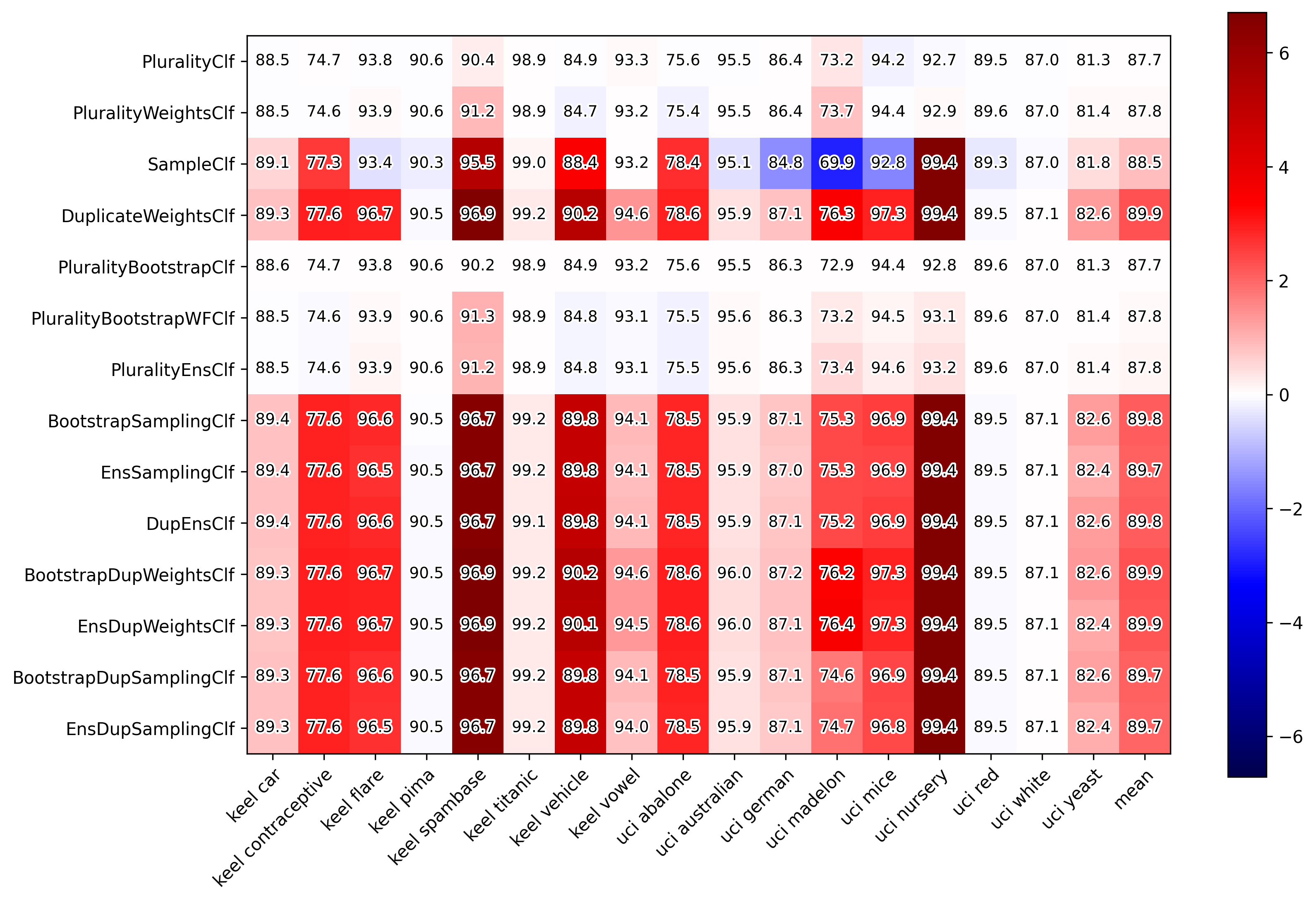}
    \caption{Base classifier: LR. Metric: AUC on $y^G$.}
    \label{fig:q1_LR_AUC}
\end{figure}

\begin{figure}[h]
    \centering
    \includegraphics[width=\textwidth, quiet]{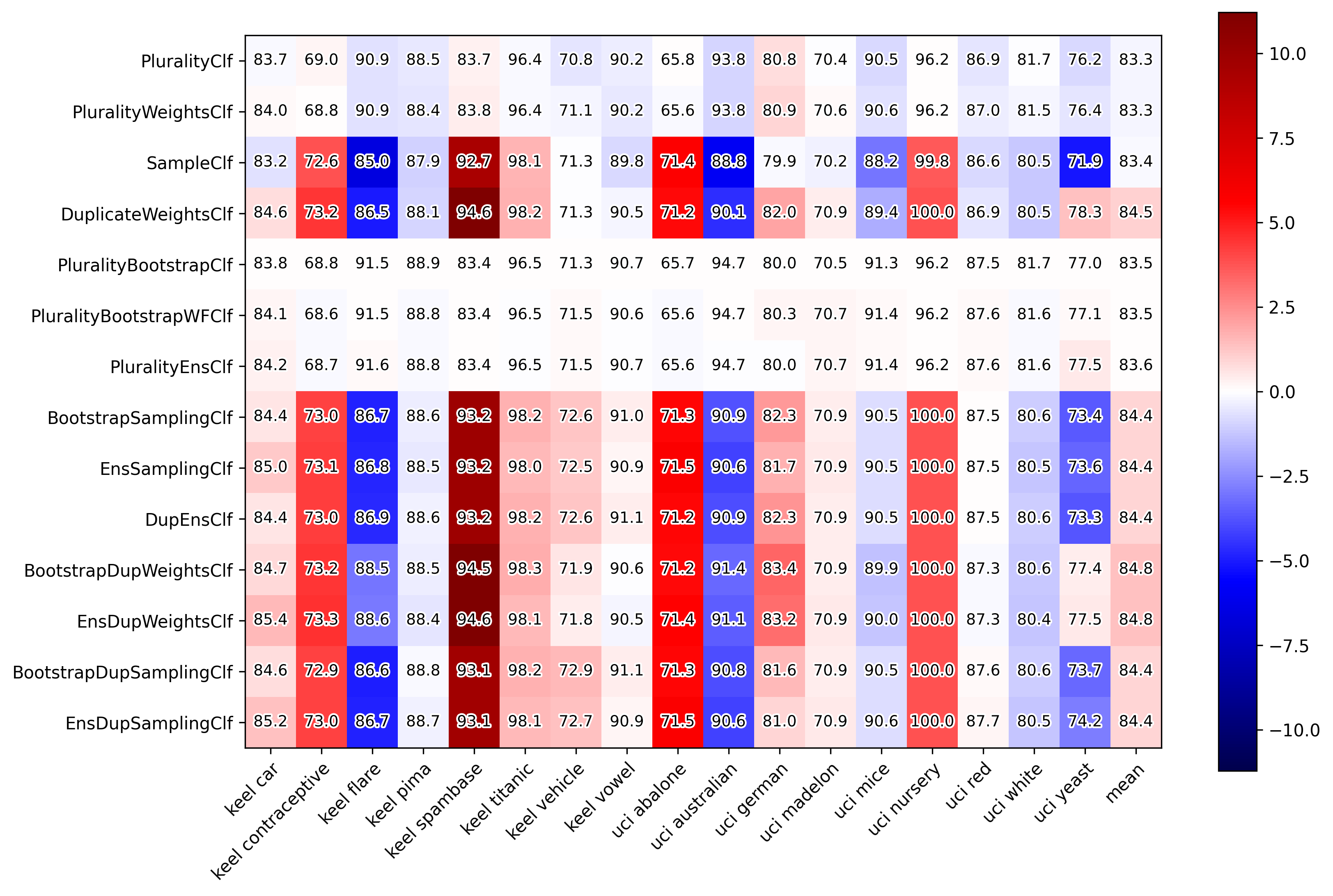}
    \caption{Base classifier: GNB. Metric: AUC on $y^G$.}
    \label{fig:q1_GNB_AUC}
\end{figure}

\begin{figure}[h]
    \centering
    \includegraphics[width=\textwidth, quiet]{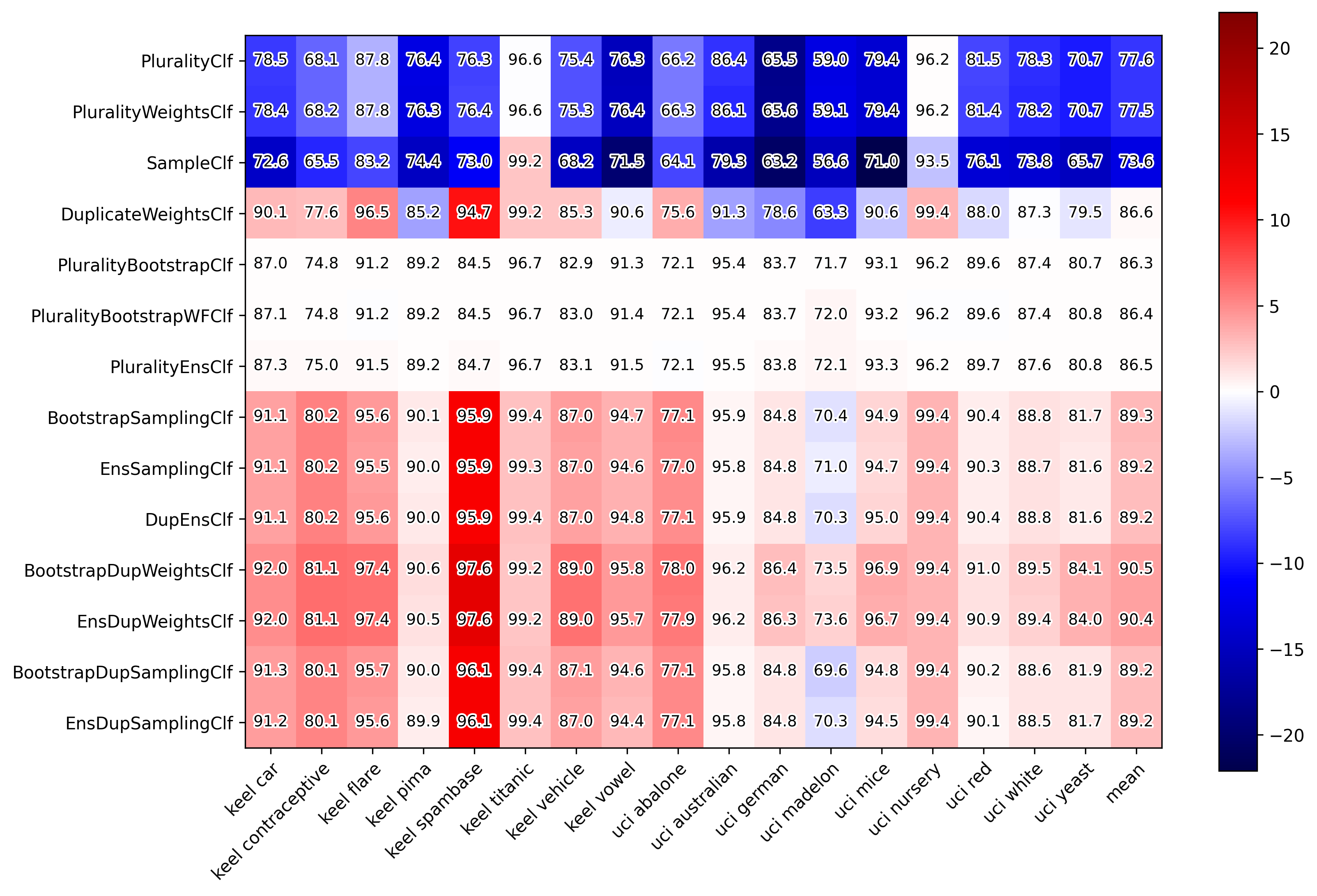}
    \caption{Base classifier: DT. Metric: AUC on $y^G$.}
    \label{fig:q1_DT_AUC}
\end{figure}

\begin{figure}[h]
    \centering
    \includegraphics[width=0.98\textwidth, quiet]{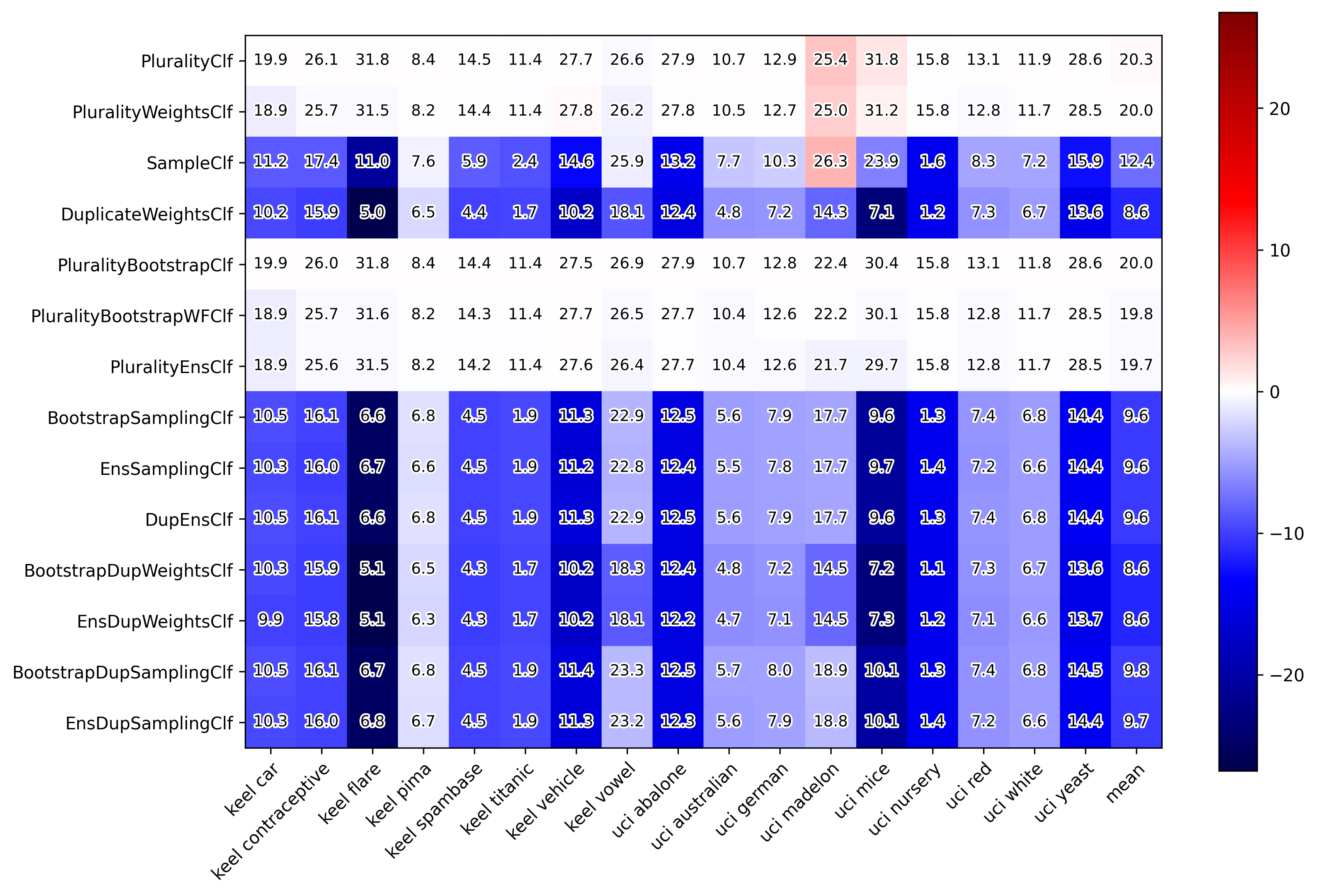}
    \caption{Base classifier: LR. Metric: $\overline{TVD}$ on $y^{PG}$.}
    \label{fig:q1_LR_TVD}
\end{figure}

\begin{figure}[h]
    \centering
    \includegraphics[width=0.98\textwidth, quiet]{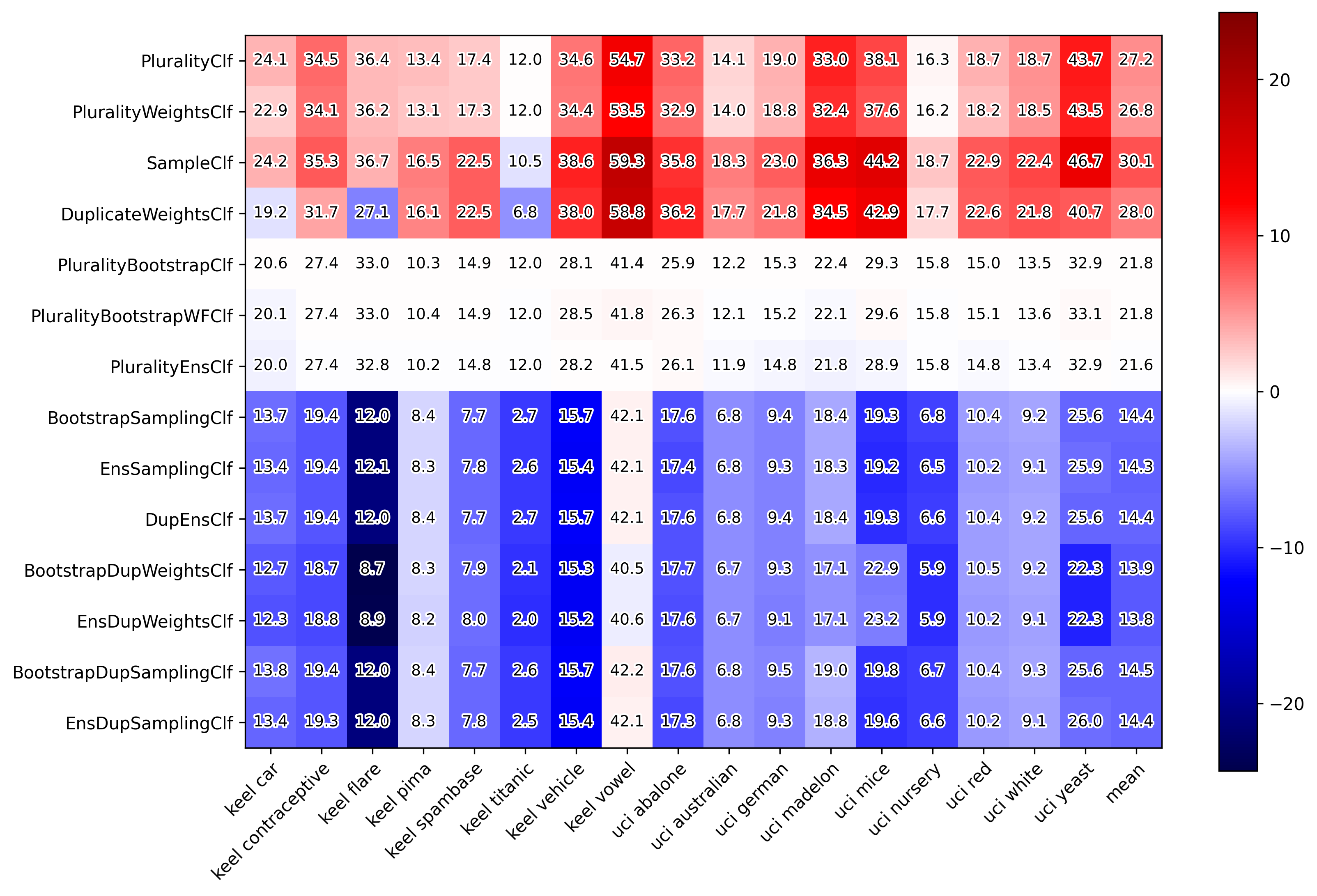}
    \caption{Base classifier: SGD. Metric: $\overline{TVD}$ on $y^{PG}$.}
    \label{fig:q1_SGD_TVD}
\end{figure}

\begin{figure}[h]
    \centering
    \includegraphics[width=0.98\textwidth, quiet]{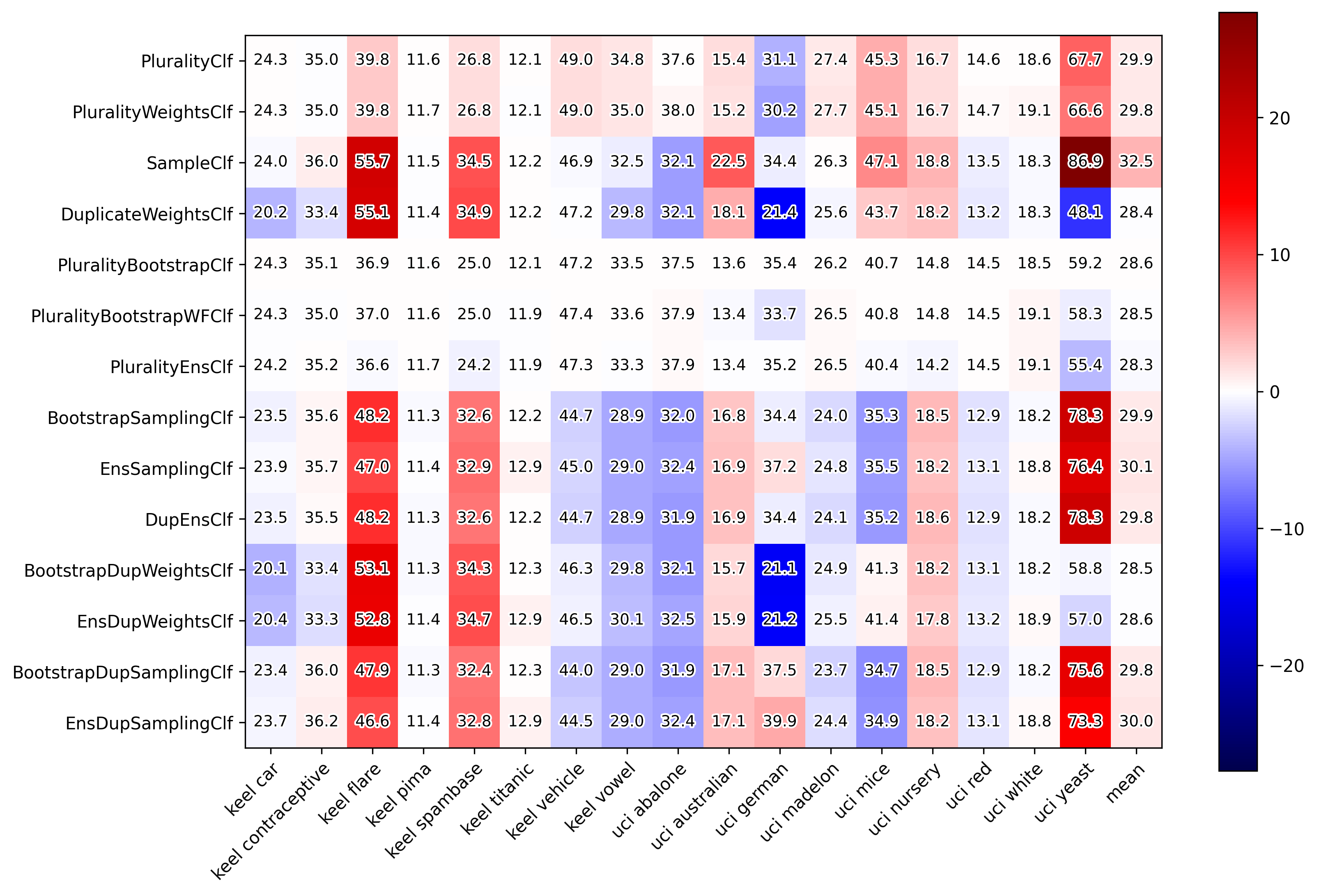}
    \caption{Base classifier: GNB. Metric: $\overline{TVD}$ on $y^{PG}$.}
    \label{fig:q1_GNB_TVD}
\end{figure}

\begin{figure}[h]
    \centering
    \includegraphics[width=0.98\textwidth, quiet]{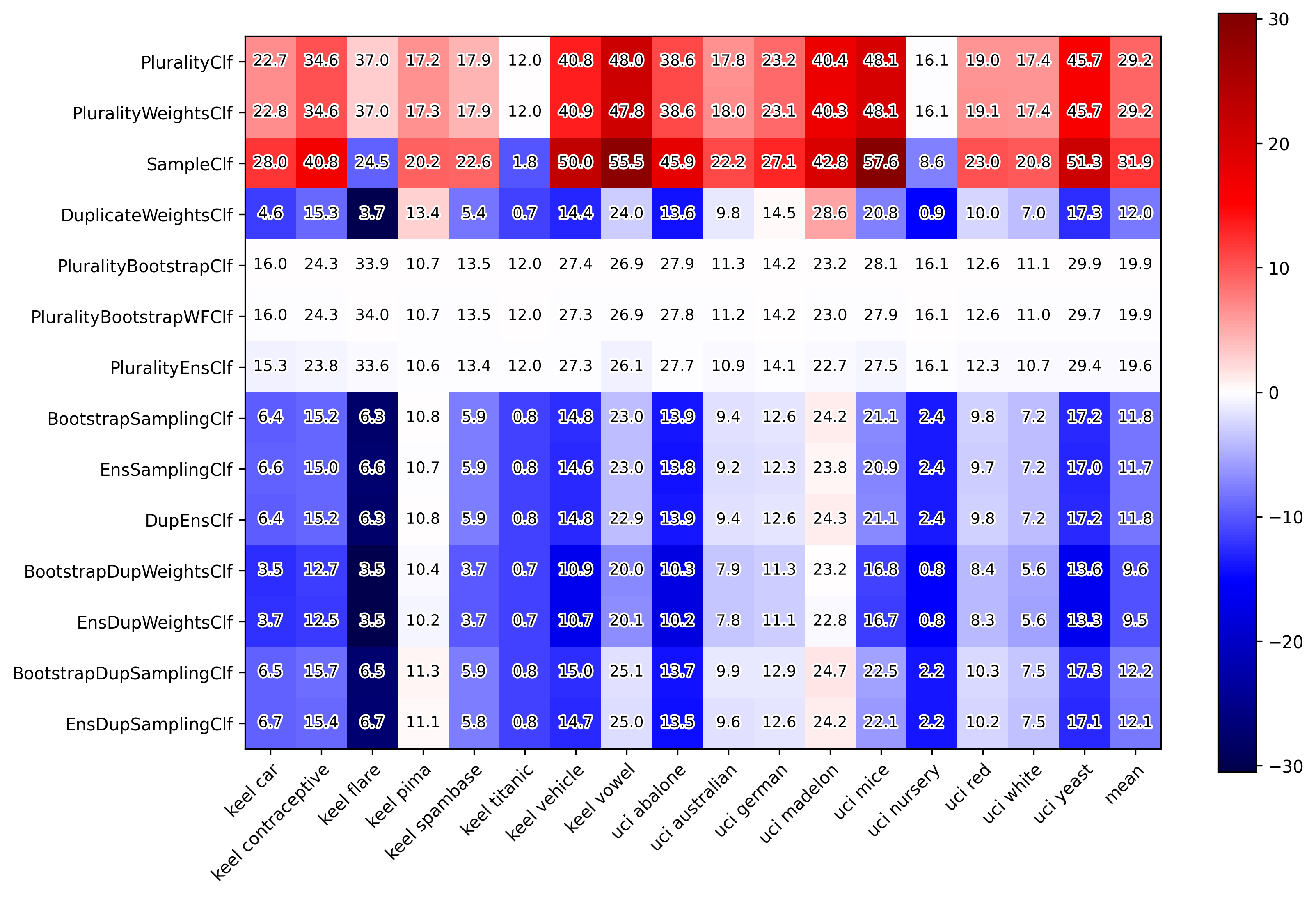}
    \caption{Base classifier: DT. Metric: $\overline{TVD}$ on $y^{PG}$.}
    \label{fig:q1_DT_TVD}
\end{figure}

\clearpage

\section{Question 1: Threshold methods} \label{appendix:threshold}

\begin{figure}[h]
    \centering
    \includegraphics[width=0.58\textwidth, quiet]{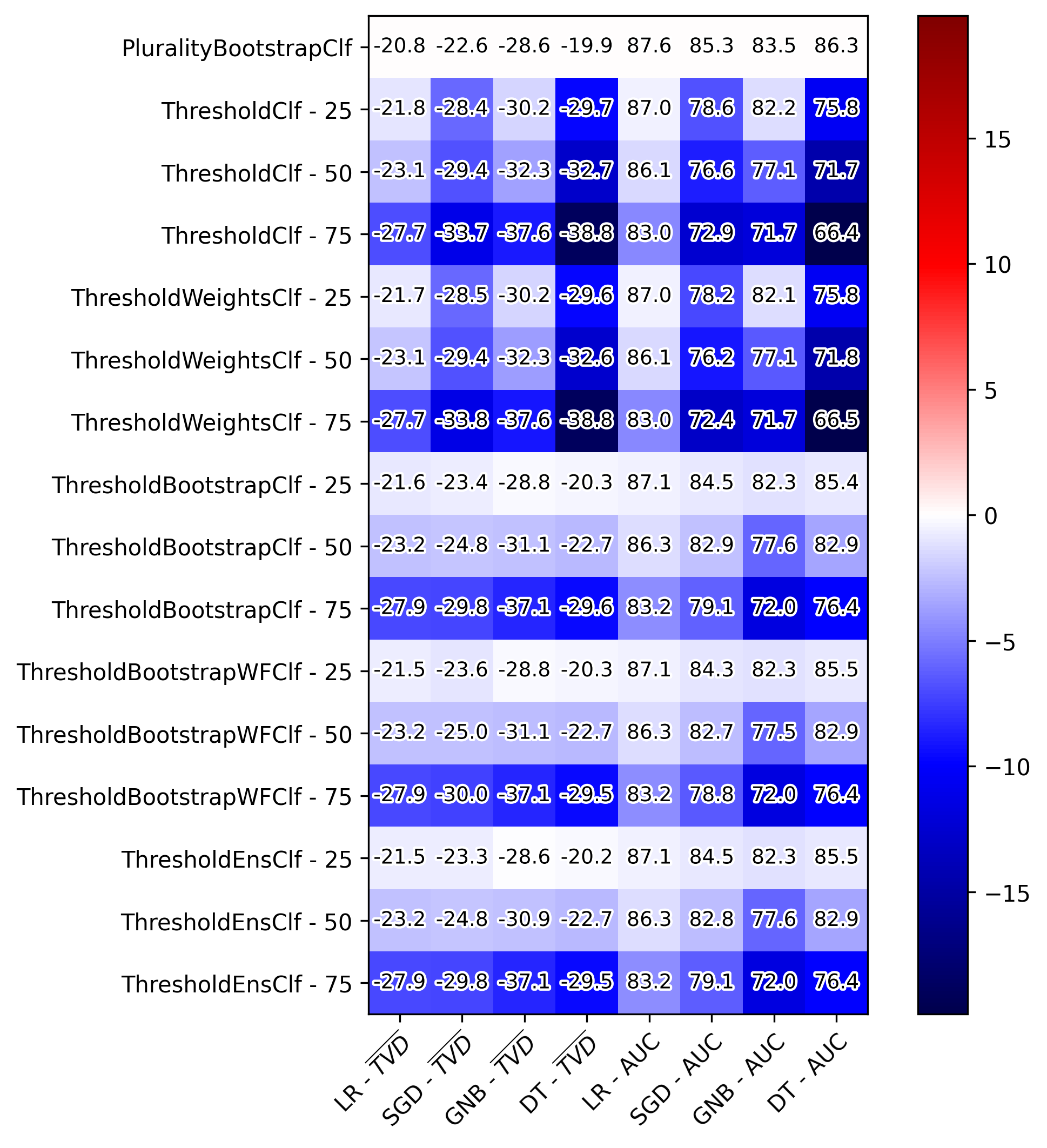}
    \caption{Heat map illustrating the performance of various threshold methods using four base classifier averaged over all datasets. Performance is measured by the AUC on $y^G$ and $\overline{TVD}$ on $y^{PG}$. The $\overline{TVD}$ values were multiplied by $-1$, to allow for easier comparison with the AUC. All values were multiplied by $100$ to enhance readability. Red cells indicate better performance, while blue cells indicate worse performance than PluralityBootstrapClf for each combination of base classifier and metric.}
    \label{fig:stat_table_threshold}
\end{figure}

\clearpage

\section{Question 2: Effect of Noise} \label{appendix:q2}

Shared description for all figures in Appendix E: The effect of six different noise types on method performance with four base classifiers, measured by AUC on the ground truth test data or $\overline{TVD}$ on the partial ground truth test data, across multiple noise levels. Noise levels range from level 0 (noiseless) to level 6 (noise strength 0.3). LR, SGD, GNB and DT were used as base classifiers. RF served as the ground truth model, with the soft labels generated at either the low or high uncertainty level

\begin{figure}[p]
    \centering
    \includegraphics[width=\textwidth, quiet]{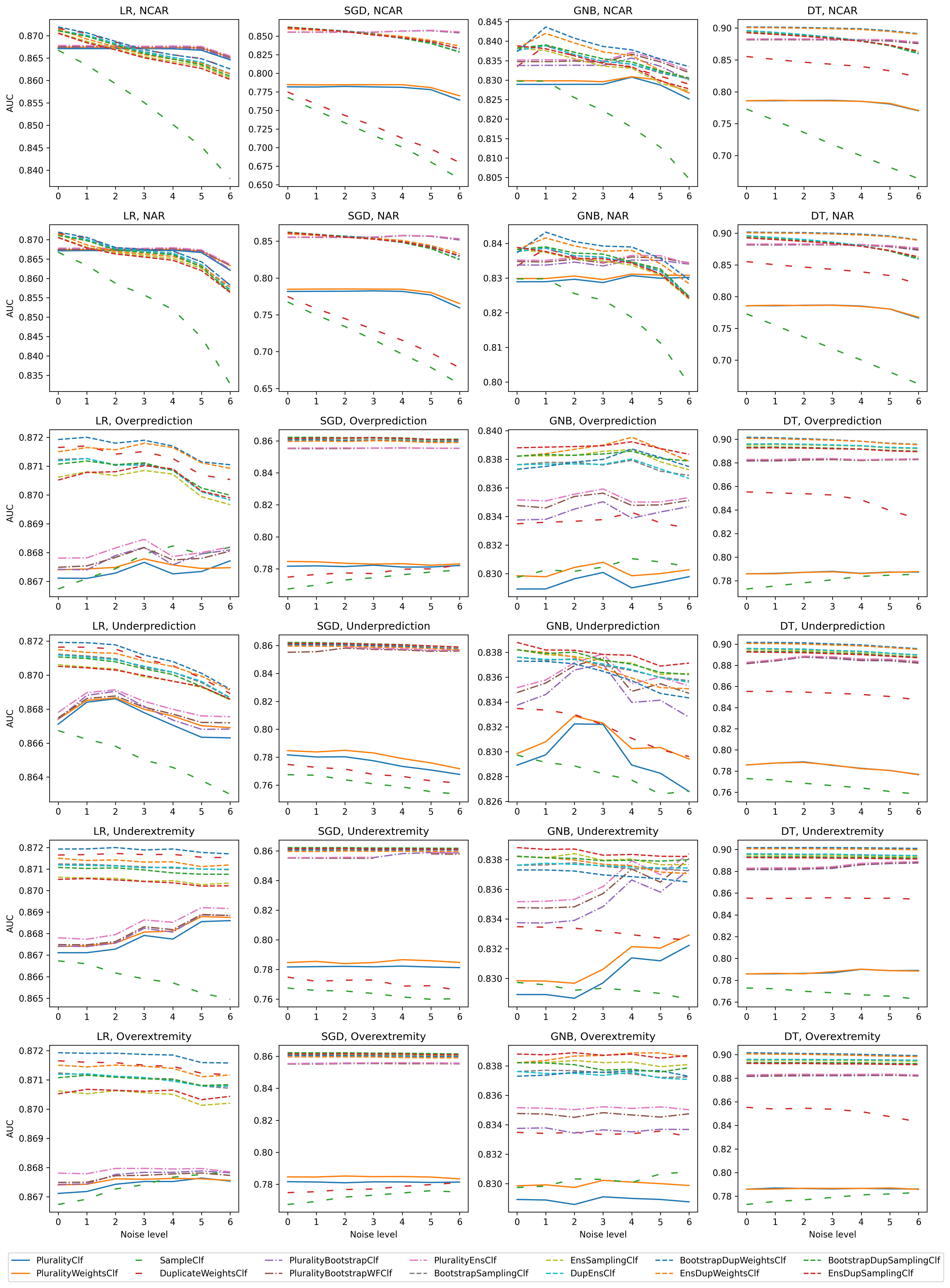}
    \caption{Test data: AUC on the ground truth. Uncertainty level: low.}
    \label{fig:q2_rf1_auc}
\end{figure}

\begin{figure}[p]
    \centering
    \includegraphics[width=\textwidth, quiet]{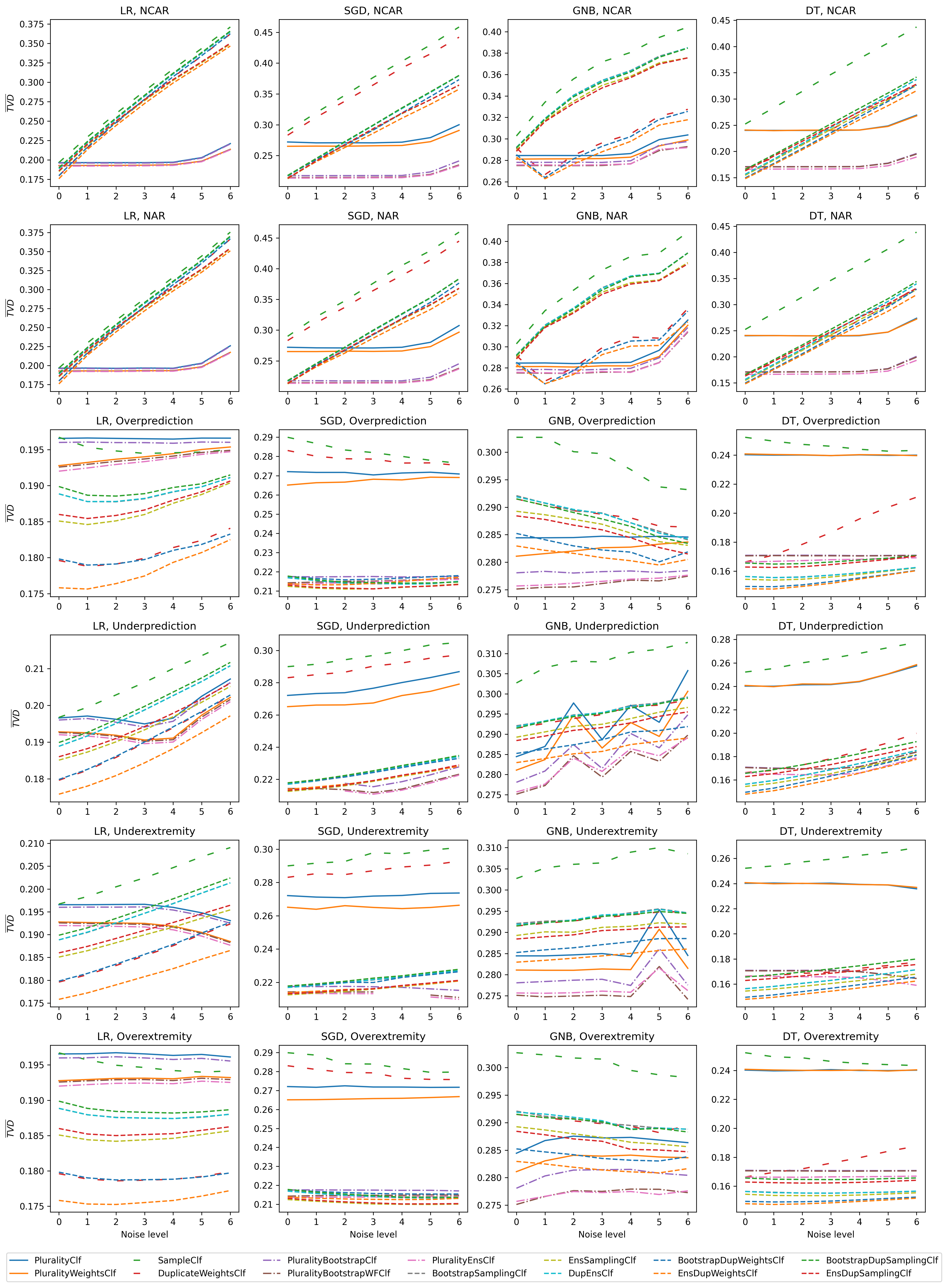}
    \caption{Test data: $\overline{TVD}$ on the partial ground truth. Uncertainty level: low.}
    \label{fig:q2_rf1_tvd}
\end{figure}

\begin{figure}[p]
    \centering
    \includegraphics[width=\textwidth, quiet]{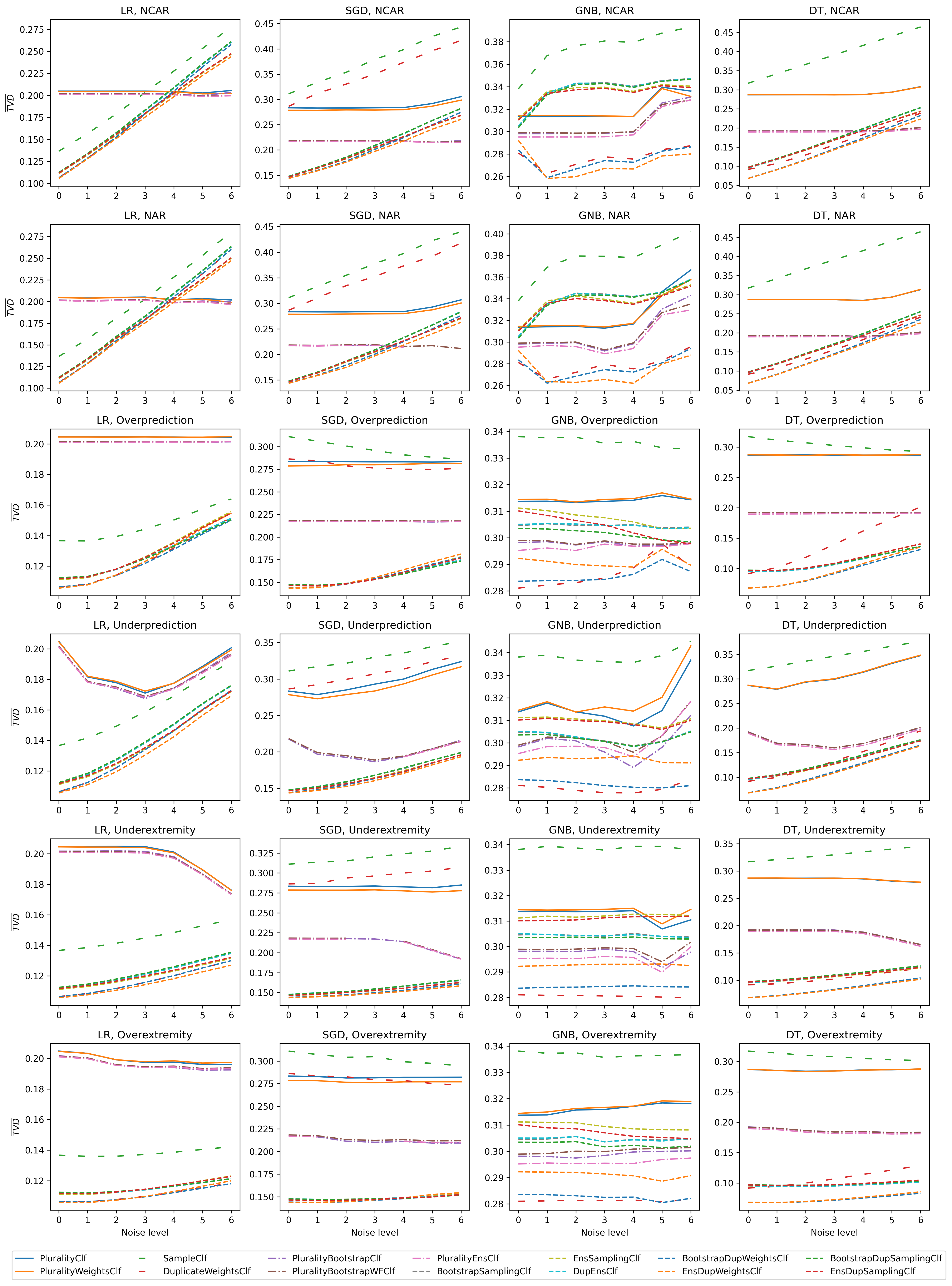}
    \caption{Test data: $\overline{TVD}$ on the partial ground truth. Uncertainty level: high.}
    \label{fig:q2_rf2_tvd}
\end{figure}

\clearpage

\section{Question 2: Relative Effect of Noise} \label{appendix:q2_base}

Shared description for all figures in Appendix F: The effect of six different noise types on method performance with four base classifiers, measured by the change in AUC or $\overline{TVD}$ relative to the noiseless baseline (level 0), on the ground truth test data. Noise levels range from level 0 (noiseless) to level 6 (noise strength 0.3). LR, SGD, GNB and DT were used as base classifiers. RF served as the ground truth model, with the soft labels generated at either the low or high uncertainty level.

\begin{figure}[p]
    \centering
    \includegraphics[width=\textwidth, quiet]{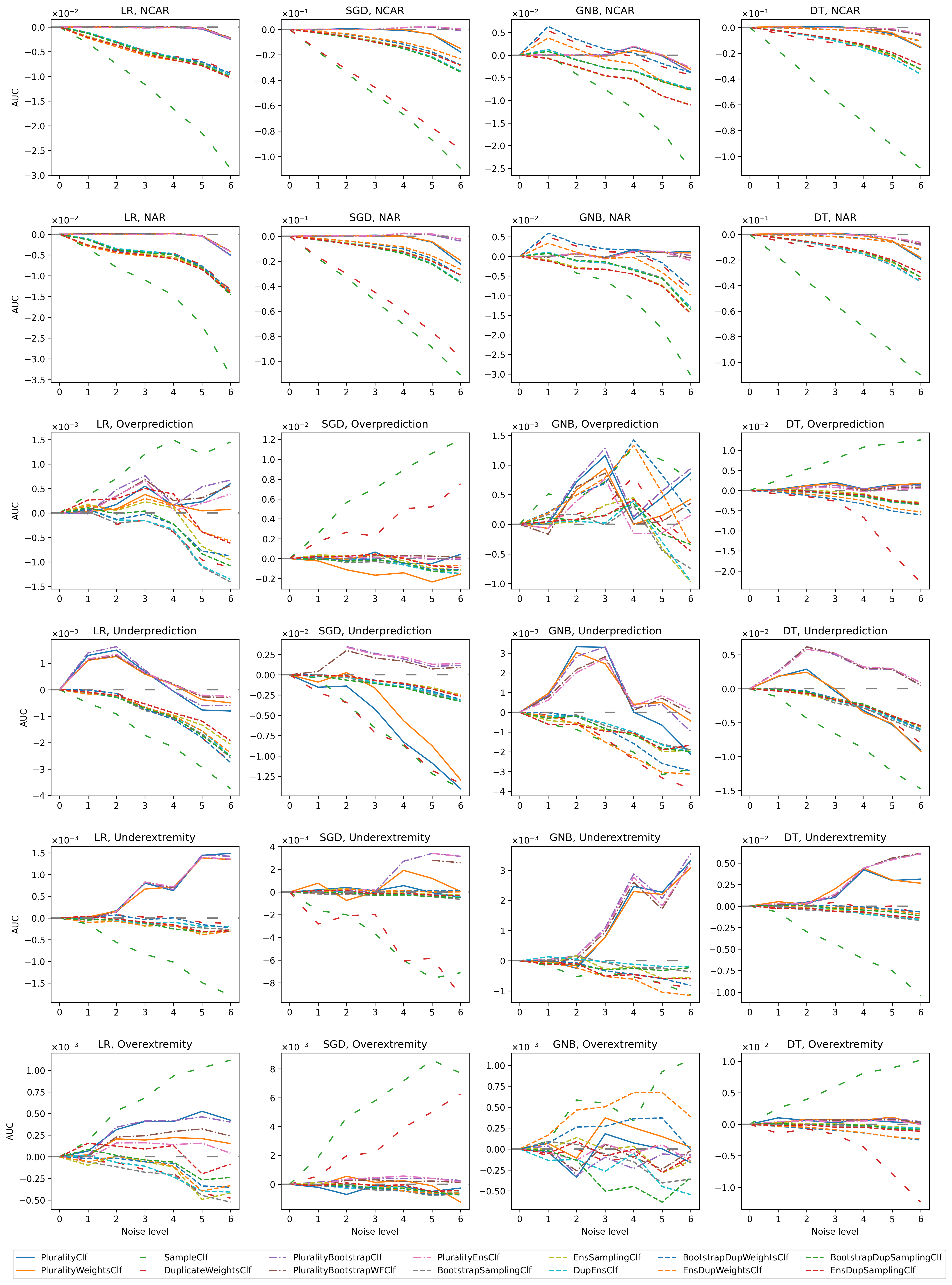}
    \caption{Test data: AUC on the ground truth. Uncertainty level: low.}
    \label{fig:q2_rf1_auc_base}
\end{figure}

\begin{figure}[p]
    \centering
    \includegraphics[width=\textwidth, quiet]{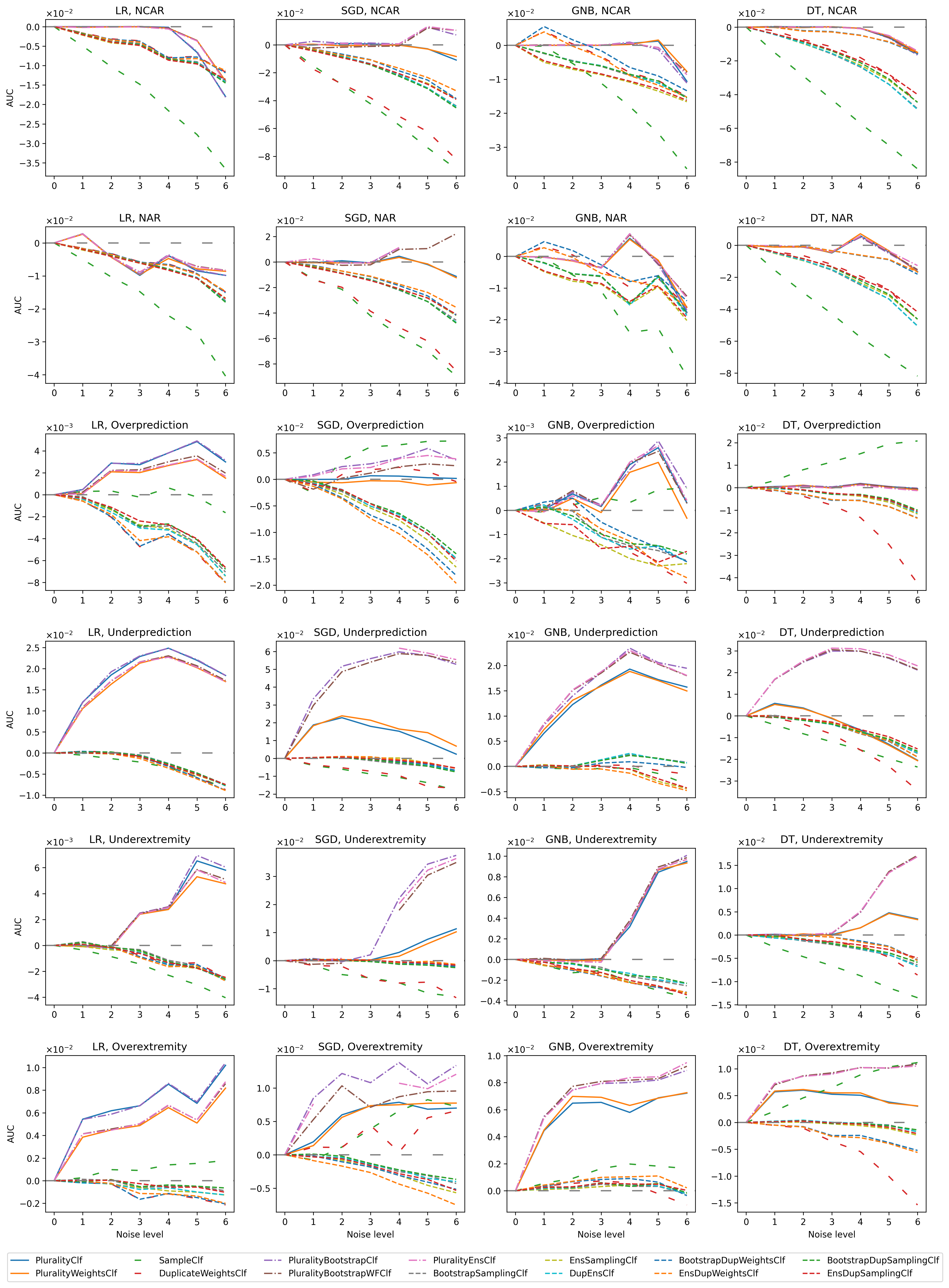}
    \caption{Test data: AUC on the ground truth. Uncertainty level: high.}
    \label{fig:q2_rf2_auc_base}
\end{figure}

\begin{figure}[p]
    \centering
    \includegraphics[width=\textwidth, quiet]{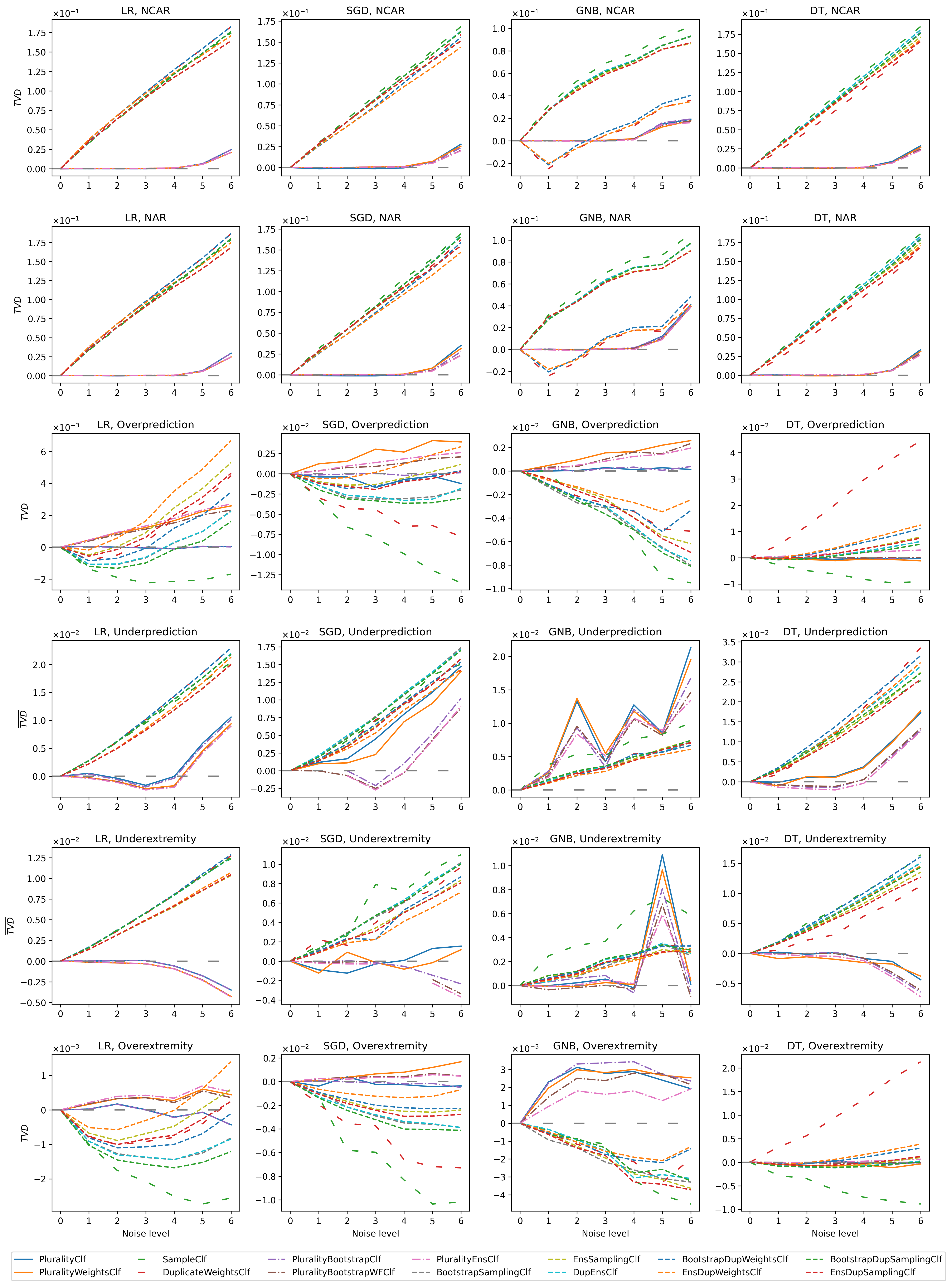}
    \caption{Test data: $\overline{TVD}$ on the partial ground truth. Uncertainty level: low.}
    \label{fig:q2_rf1_tvd_base}
\end{figure}

\begin{figure}[p]
    \centering
    \includegraphics[width=\textwidth, quiet]{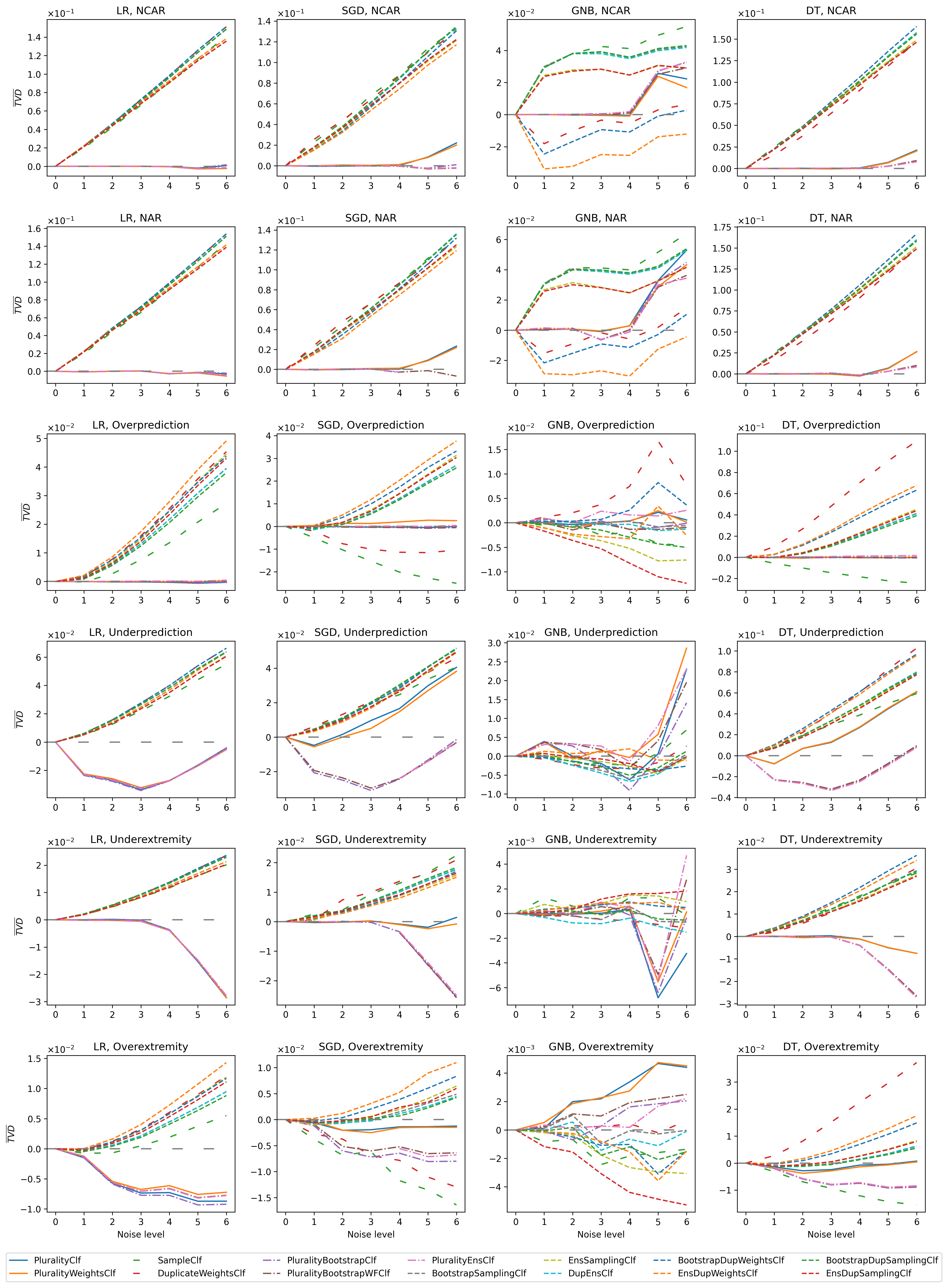}
    \caption{Test data: $\overline{TVD}$ on the partial ground truth. Uncertainty level: high.}
    \label{fig:q2_rf2_tvd_base}
\end{figure}

\clearpage

\section{Real-World Data}\label{appendix:rwd}

\begin{figure}[h]
    \centering
    \includegraphics[width=\textwidth, quiet]{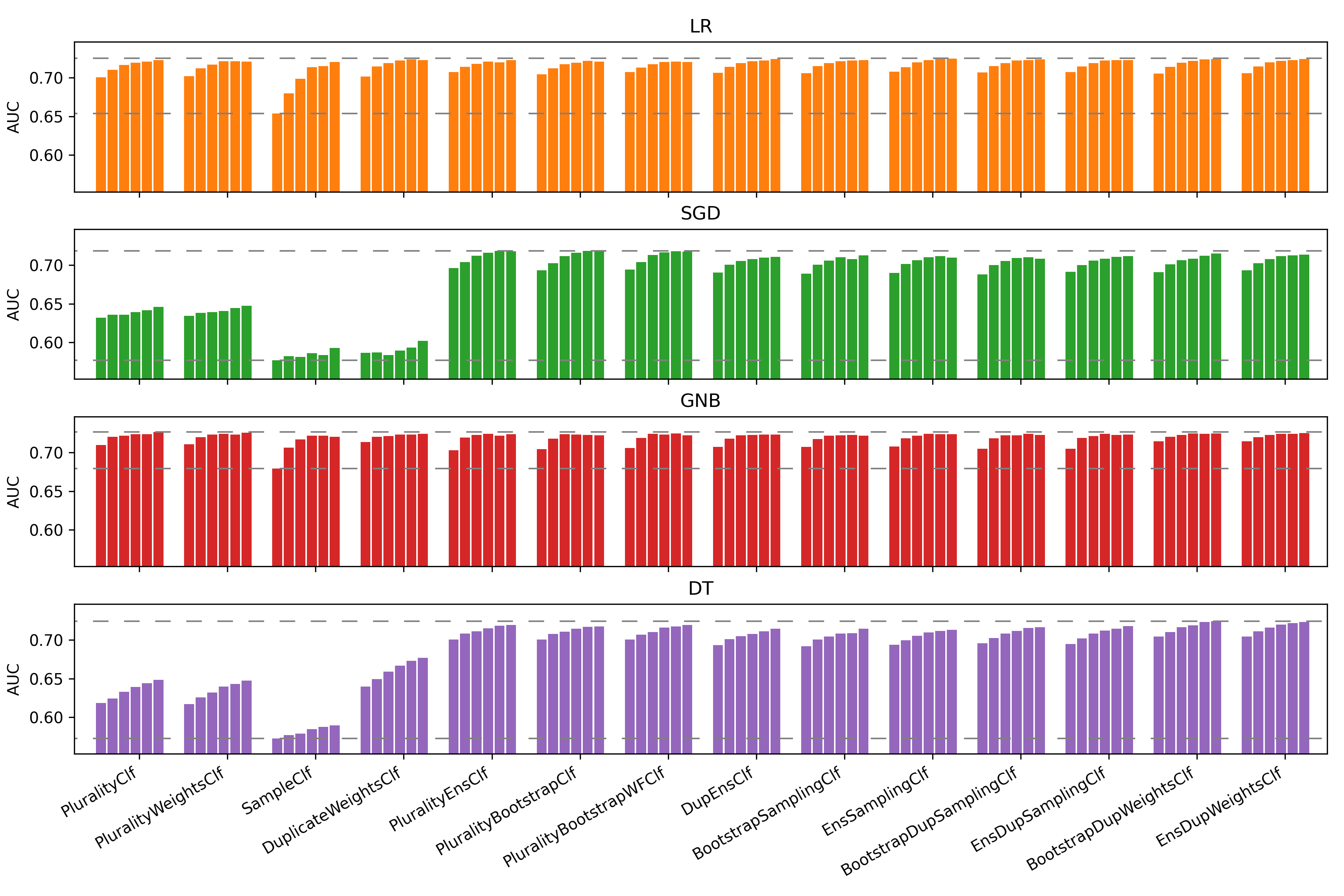}
    \caption{Method performance on the UrinCheck dataset, measured by AUC. The labels used for the test set have been sampled from the soft-labelled data. For each method the individual bars represent the fraction of the total data that was used as training data in that experiment, increasing from left to right: $\{0.05,0.1,0.2,0.4,0.6,0.8\}$}
    \label{fig:rw_auc_samp}
\end{figure}

\end{appendices}

\bibliography{SLL_references}

\begin{thebibliography}{42}
\providecommand{\natexlab}[1]{#1}
\providecommand{\url}[1]{{#1}}
\providecommand{\urlprefix}{URL }
\providecommand{\doi}[1]{\url{https://doi.org/#1}}
\providecommand{\eprint}[2][]{\url{#2}}
 \bibcommenthead

\bibitem[{Alcal{\'{a}}-Fdez et~al(2011)Alcal{\'{a}}-Fdez, Fern{\'{a}}ndez,
  Luengo, Derrac, Garc{\'{i}}a, S{\'{a}}nchez, and Herrera}]{Alcala-Fdez2011}
Alcal{\'{a}}-Fdez J, Fern{\'{a}}ndez A, Luengo J, et~al (2011) Keel data-mining
  software tool: Data set repository, integration of algorithms and
  experimental analysis framework. Journal of Multiple-Valued Logic and Soft
  Computing 17(2-3):255--287

\bibitem[{Berthon et~al(2021)Berthon, Han, Niu, Liu, and
  Sugiyama}]{berthon2021confidence}
Berthon A, Han B, Niu G, et~al (2021) Confidence scores make instance-dependent
  label-noise learning possible. In: International Conference on Machine
  Learning, PMLR, pp 825--836

\bibitem[{Breiman(1996)}]{breiman1996bagging}
Breiman L (1996) Bagging predictors. Machine learning 24:123--140

\bibitem[{Brown et~al(2005)Brown, Wyatt, Harris, and Yao}]{brown2005diversity}
Brown G, Wyatt J, Harris R, et~al (2005) Diversity creation methods: a survey
  and categorisation. Information fusion 6(1):5--20

\bibitem[{C{\^o}me et~al(2009)C{\^o}me, Oukhellou, Denoeux, and
  Aknin}]{come2009learning}
C{\^o}me E, Oukhellou L, Denoeux T, et~al (2009) Learning from partially
  supervised data using mixture models and belief functions. Pattern
  recognition 42(3):334--348

\bibitem[{Cortez et~al(2009)Cortez, Cerdeira, Almeida, Matos, and
  Reis}]{cortez2009modeling}
Cortez P, Cerdeira A, Almeida F, et~al (2009) Modeling wine preferences by data
  mining from physicochemical properties. Decision support systems
  47(4):547--553

\bibitem[{Dawid and Skene(1979)}]{dawid1979maximum}
Dawid AP, Skene AM (1979) Maximum likelihood estimation of observer error-rates
  using the em algorithm. Journal of the Royal Statistical Society: Series C
  (Applied Statistics) 28(1):20--28

\bibitem[{Den{\oe}ux and Zouhal(2001)}]{denoeux2001handling}
Den{\oe}ux T, Zouhal LM (2001) Handling possibilistic labels in pattern
  classification using evidential reasoning. Fuzzy sets and systems
  122(3):409--424

\bibitem[{Dua and Graff(2017)}]{Dua:2019}
Dua D, Graff C (2017) {UCI} machine learning repository.
  \urlprefix\url{http://archive.ics.uci.edu/ml}

\bibitem[{Fornaciari et~al(2021)Fornaciari, Uma, Paun, Plank, Hovy, and
  Poesio}]{fornaciari2021beyond}
Fornaciari T, Uma A, Paun S, et~al (2021) Beyond black \& white: Leveraging
  annotator disagreement via soft-label multi-task learning. In: 2021
  Conference of the North American Chapter of the Association for Computational
  Linguistics: Human Language Technologies, Association for Computational
  Linguistics

\bibitem[{Fr{\'e}nay and Verleysen(2013)}]{frenay2013classification}
Fr{\'e}nay B, Verleysen M (2013) Classification in the presence of label noise:
  a survey. IEEE transactions on neural networks and learning systems
  25(5):845--869

\bibitem[{Gao et~al(2017)Gao, Xing, Xie, Wu, and Geng}]{gao2017deep}
Gao BB, Xing C, Xie CW, et~al (2017) Deep label distribution learning with
  label ambiguity. IEEE Transactions on Image Processing 26(6):2825--2838

\bibitem[{Garc{\'\i}a et~al(2010)Garc{\'\i}a, Fern{\'a}ndez, Luengo, and
  Herrera}]{garcia2010advanced}
Garc{\'\i}a S, Fern{\'a}ndez A, Luengo J, et~al (2010) Advanced nonparametric
  tests for multiple comparisons in the design of experiments in computational
  intelligence and data mining: Experimental analysis of power. Information
  sciences 180(10):2044--2064

\bibitem[{Geng(2016)}]{geng2016label}
Geng X (2016) Label distribution learning. IEEE Transactions on Knowledge and
  Data Engineering 28(7):1734--1748

\bibitem[{Griffin and Brenner(2004)}]{griffin2004perspectives}
Griffin D, Brenner L (2004) Perspectives on probability judgment calibration.
  Blackwell handbook of judgment and decision making 199:158--177

\bibitem[{Griffin and Tversky(1992)}]{griffin1992weighing}
Griffin D, Tversky A (1992) The weighing of evidence and the determinants of
  confidence. Cognitive psychology 24(3):411--435

\bibitem[{Gui et~al(2015)Gui, Lu, Xu, Li, and Wei}]{gui2015novel}
Gui L, Lu Q, Xu R, et~al (2015) A novel class noise estimation method and
  application in classification. In: Proceedings of the 24th ACM International
  on Conference on Information and Knowledge Management, pp 1081--1090

\bibitem[{Jin and Ghahramani(2002)}]{jin2002learning}
Jin R, Ghahramani Z (2002) Learning with multiple labels. Advances in neural
  information processing systems 15

\bibitem[{Lichtenstein et~al(1977)Lichtenstein, Fischhoff, and
  Phillips}]{lichtenstein1977calibration}
Lichtenstein S, Fischhoff B, Phillips LD (1977) Calibration of probabilities:
  The state of the art. In: Decision Making and Change in Human Affairs:
  Proceedings of the Fifth Research Conference on Subjective Probability,
  Utility, and Decision Making, Darmstadt, 1--4 September, 1975, Springer, pp
  275--324

\bibitem[{Maci{\`{a}} and Bernad{\'{o}}-Mansilla(2014)}]{Macia2014}
Maci{\`{a}} N, Bernad{\'{o}}-Mansilla E (2014) Towards {UCI+}: A mindful
  repository design. Information Sciences 261:237--262.
  \doi{10.1016/j.ins.2013.08.059}

\bibitem[{Nguyen et~al(2011)Nguyen, Valizadegan, and
  Hauskrecht}]{nguyen2011learning}
Nguyen Q, Valizadegan H, Hauskrecht M (2011) Learning classification with
  auxiliary probabilistic information. In: 2011 IEEE 11th International
  Conference on Data Mining, IEEE, pp 477--486

\bibitem[{Nguyen et~al(2014)Nguyen, Valizadegan, and
  Hauskrecht}]{nguyen2014learning}
Nguyen Q, Valizadegan H, Hauskrecht M (2014) Learning classification models
  with soft-label information. Journal of the American Medical Informatics
  Association 21(3):501--508

\bibitem[{Oyama et~al(2013)Oyama, Baba, Sakurai, and
  Kashima}]{oyama2013accurate}
Oyama S, Baba Y, Sakurai Y, et~al (2013) Accurate integration of crowdsourced
  labels using workers' self-reported confidence scores. In: Twenty-Third
  International Joint Conference on Artificial Intelligence

\bibitem[{Pedregosa et~al(2011)Pedregosa, Varoquaux, Gramfort, Michel, Thirion,
  Grisel, Blondel, Prettenhofer, Weiss, Dubourg, Vanderplas, Passos,
  Cournapeau, Brucher, Perrot, and Duchesnay}]{scikit-learn}
Pedregosa F, Varoquaux G, Gramfort A, et~al (2011) Scikit-learn: Machine
  learning in {P}ython. Journal of Machine Learning Research 12:2825--2830

\bibitem[{Peng et~al(2014)Peng, Wong, and Yu}]{peng2014learning}
Peng P, Wong RCW, Yu PS (2014) Learning on probabilistic labels. In:
  Proceedings of the 2014 SIAM International Conference on Data Mining, SIAM,
  pp 307--315

\bibitem[{Peterson et~al(2019)Peterson, Battleday, Griffiths, and
  Russakovsky}]{peterson2019human}
Peterson JC, Battleday RM, Griffiths TL, et~al (2019) Human uncertainty makes
  classification more robust. In: Proceedings of the IEEE/CVF International
  Conference on Computer Vision, pp 9617--9626

\bibitem[{Raykar et~al(2010)Raykar, Yu, Zhao, Valadez, Florin, Bogoni, and
  Moy}]{raykar2010learning}
Raykar VC, Yu S, Zhao LH, et~al (2010) Learning from crowds. Journal of machine
  learning research 11(4)

\bibitem[{Reamaroon et~al(2018)Reamaroon, Sjoding, Lin, Iwashyna, and
  Najarian}]{reamaroon2018accounting}
Reamaroon N, Sjoding MW, Lin K, et~al (2018) Accounting for label uncertainty
  in machine learning for detection of acute respiratory distress syndrome.
  IEEE journal of biomedical and health informatics 23(1):407--415

\bibitem[{Sheng(2011)}]{sheng2011simple}
Sheng VS (2011) Simple multiple noisy label utilization strategies. In: 2011
  IEEE 11th International Conference on Data Mining, IEEE, pp 635--644

\bibitem[{Sheng et~al(2008)Sheng, Provost, and Ipeirotis}]{sheng2008get}
Sheng VS, Provost F, Ipeirotis PG (2008) Get another label? improving data
  quality and data mining using multiple, noisy labelers. In: Proceedings of
  the 14th ACM SIGKDD international conference on Knowledge discovery and data
  mining, pp 614--622

\bibitem[{Song et~al(2018)Song, Wang, Gao, and An}]{song2018active}
Song J, Wang H, Gao Y, et~al (2018) Active learning with confidence-based
  answers for crowdsourcing labeling tasks. Knowledge-Based Systems
  159:244--258

\bibitem[{Tversky and Kahneman(1974)}]{tversky1974judgment}
Tversky A, Kahneman D (1974) Judgment under uncertainty: Heuristics and biases:
  Biases in judgments reveal some heuristics of thinking under uncertainty.
  science 185(4157):1124--1131

\bibitem[{de~Vries and Thierens(2021)}]{de2021reliable}
de~Vries S, Thierens D (2021) A reliable ensemble based approach to
  semi-supervised learning. Knowledge-Based Systems 215:106738

\bibitem[{de~Vries and Thierens(2024)}]{de2023generating}
de~Vries S, Thierens D (2024) Generating the ground truth: Synthetic data for
  soft label and label noise research. arXiv preprint arXiv:230904318

\bibitem[{de~Vries et~al(2022)de~Vries, Ten~Doesschate, Tott{\'e}, Heutz,
  Loeffen, Oosterheert, Thierens, and Boel}]{de2022semi}
de~Vries S, Ten~Doesschate T, Tott{\'e} JE, et~al (2022) A semi-supervised
  decision support system to facilitate antibiotic stewardship for urinary
  tract infections. Computers in Biology and Medicine 146:105621

\bibitem[{Xue and Hauskrecht(2017)}]{xue2017efficient}
Xue Y, Hauskrecht M (2017) Efficient learning of classification models from
  soft-label information by binning and ranking. In: The Thirtieth
  International Flairs Conference

\bibitem[{Yang and Thompson(2010)}]{yang2010nurses}
Yang H, Thompson C (2010) Nurses’ risk assessment judgements: A confidence
  calibration study. Journal of Advanced Nursing 66(12):2751--2760

\bibitem[{Zadrozny et~al(2003)Zadrozny, Langford, and Abe}]{zadrozny2003cost}
Zadrozny B, Langford J, Abe N (2003) Cost-sensitive learning by
  cost-proportionate example weighting. In: Third IEEE international conference
  on data mining, IEEE, pp 435--442

\bibitem[{Zhang(2022)}]{zhang2022knowledge}
Zhang J (2022) Knowledge learning with crowdsourcing: A brief review and
  systematic perspective. IEEE/CAA Journal of Automatica Sinica 9(5):749--762

\bibitem[{Zhang et~al(2016)Zhang, Wu, and Sheng}]{zhang2016learning}
Zhang J, Wu X, Sheng VS (2016) Learning from crowdsourced labeled data: a
  survey. Artificial Intelligence Review 46(4):543--576

\bibitem[{Zhang et~al(2018)Zhang, Wu, and Sheng}]{zhang2018ensemble}
Zhang J, Wu M, Sheng VS (2018) Ensemble learning from crowds. IEEE Transactions
  on Knowledge and Data Engineering 31(8):1506--1519

\bibitem[{Zhu and Wu(2004)}]{zhu2004class}
Zhu X, Wu X (2004) Class noise vs. attribute noise: A quantitative study.
  Artificial intelligence review 22:177--210

\end{thebibliography}

\end{document}